\pdfoutput=1

\documentclass[11pt]{article}

\PassOptionsToPackage{table}{xcolor}
\usepackage[]{acl}

\usepackage{times}
\usepackage{latexsym}

\usepackage{pgfplots}
\pgfplotsset{compat=1.18}

\usepackage[T1]{fontenc}

\usepackage[utf8]{inputenc}

\usepackage{microtype}

\usepackage{inconsolata}
\usepackage{enumitem}
\usepackage[most]{tcolorbox}
\usepackage{bbding}
\usepackage{adjustbox}
\usepackage{multirow}
\usepackage{caption}
\usepackage{amsmath}
\usepackage{amssymb}
\usepackage{amsfonts}
\usepackage{xcolor}
\usepackage{graphicx}
\usepackage{booktabs}
\usepackage{mathtools}
\usepackage[title]{appendix}
\usepackage{listings}

\newcommand{\squishlist}{
 \begin{list}{$\bullet$}
  { \setlength{\itemsep}{0pt}
     \setlength{\parsep}{3pt}
     \setlength{\topsep}{3pt}
     \setlength{\partopsep}{0pt}
     \setlength{\leftmargin}{1.5em}
     \setlength{\labelwidth}{1em}
     \setlength{\labelsep}{0.5em} } }
\newcommand{\squishend}{
     \end{list}}

 \newcommand\blfootnote[1]{%
  \begingroup
  \renewcommand\thefootnote{}\footnote{#1}%
  \addtocounter{footnote}{-1}%
  \endgroup
}

\title{Investigating LLM Capabilities on Long Context Comprehension for Clinical Question Answering
}

\author{
\textbf{Feras AlMannaa$^{1*}$, Talia Tseriotou$^{2*}$, Jenny Chim$^{2}$, Maria Liakata$^{2,3}$}\\
       $^1$Istanbul Aydın University\\
       $^2$Queen Mary University of London \\$^3$The Alan Turing Institute\\
      \small \tt  \{firas.almanna3\}@gmail.com,\{t.tseriotou,m.liakata\}@qmul.ac.uk}

\begin{document}
\maketitle


\blfootnote{* Indicates equal contribution.}

\renewcommand*{\thefootnote}{\arabic{footnote}}

\begin{abstract}

Despite the impressive reported performance of Large Language Models (LLMs) on medical Question-Answering (QA), evaluation still overrelies on multiple-choice and extractive settings primarily assessing textbook knowledge, with limited work on open-ended QA over patient records. We present a comprehensive study investigating LLM long-context (LC) comprehension on longitudinal, patient-centric Electronic Health Records (EHRs) across multiple datasets, models and metrics, covering extractive, multiple-choice, and open-ended generative QA under different context formulation and Retrieval Augmented Generation (RAG) settings. We show that medically fine-tuned models offer no measurable benefit over base counterparts, smaller models exhibit stronger memorization, and longitudinal supplementary notes aid generative QA. Reasoning models gain on complex multi-note QA under LLM Correctness but not stricter reference-aligned metrics, while hybrid RAG with reranking outperforms Full-Context (FC) prompting. 
A question-category analysis and multi-faceted metric evaluation further reveal where RAG and FC excel and struggle, with RAG falling behind in longitudinal reasoning, thus uncovering systematic failures pointing to the need for reasoning-appropriate metrics.



\end{abstract}

\section{Introduction}
Large Language Models (LLMs) \citep{Achiam2023GPT4TR, Dubey2024TheL3, Yang2024Qwen25TR} perform impressively on medical tasks \citep{Borgeaud2021ImprovingLM, nori2023capabilities}, achieving superhuman scores on United States Medical Licensing Examination
(USMLE)-style exam question-answering (QA)~\citep{chen2023meditron, Tang2023MedAgentsLL, Pal2024GeminiGT, singhal2025toward}. 
Yet evaluation is mostly based on multiple-choice QA (MCQA) \citep{lievin2024can, xiong2024benchmarking, chen2024huatuogpt, singhal2025toward} which doesn't reflect performance in complex tasks \cite{arias2025automatic}, such as open-ended medical QA \citep{sandeep2024few}. 
Furthermore, medical board exams designed to assess professional knowledge and decision-making rely primarily on textbook knowledge which is likely available during pre-training~\citep{chen2025benchmarking}. 
With clinicians facing increasing workloads \citep{martinez2024patient}, LLM-based QA on patient EHRs offers promise, yet existing deployment studies rely on shallow, category-triggered context and narrow clinical evaluation \citep{Garcia2024, Bootsma2024AI}. Systematic assessment on expert-curated EHRs would signal how well models handle real-world clinical complexity, yet research focuses largely on dataset creation \citep{Yang2022ALL} or specialized models \citep{Fleming2023MedAlignAC} rather than on assessing LLMs on realistic long-context clinical settings. Related work has examined LLMs for diagnostic prediction over EHRs \citep{pan2025Integrating, albassam2025leveraging}, but is limited 
to specific diagnoses or single datasets and to short-form accuracy-based 
evaluation.



\begin{figure*}
\centering
\includegraphics[width=0.95\linewidth]{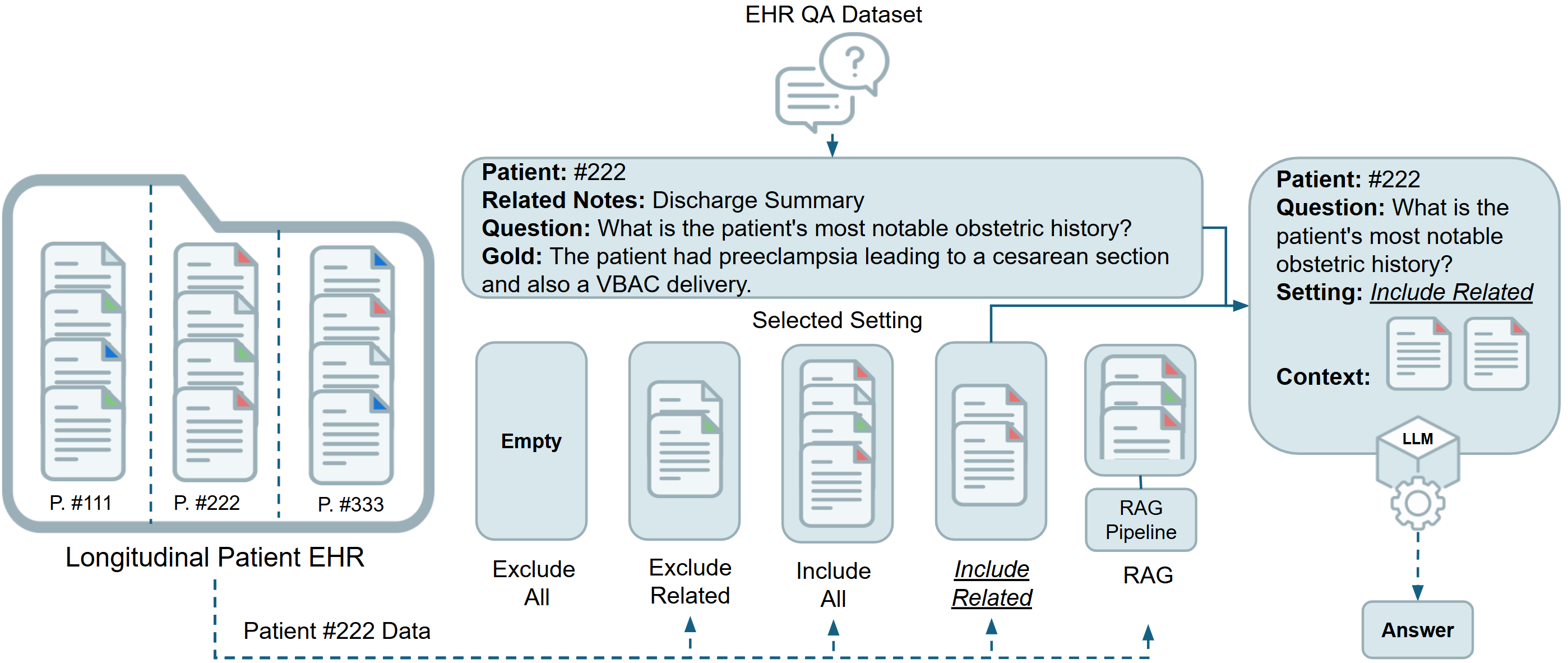}
\caption{Overview of our Approach. Given a patient question grounded in EHRs, the most related note (per original dataset) and an expert-verified gold answer (used only for evaluation), we augment the relevant note with the patient's full longitudinal EHR history from MIMIC (\S \ref{sec:contextform}). We evaluate four prompting settings: (a) \textit{Include Related}: only the most related note(s), (b) \textit{Include All}: all longitudinal notes, (c) \textit{Exclude Related}: supplementary notes only, (d) \textit{Exclude All}: no notes. We also examine two temporally ordered RAG strategies over all notes: (i) \textit{RAG}: retrieved 512-token chunks, (ii) \textit{RAG HIR}: hierarchical parent note inclusion by substituting retrieved chunks with their full parent note (\S \ref{sec:evalframework}, \ref{sec:ragsetup}). Evaluation spans
two extractive and two open-ended generative QA datasets.}
\label{fig:intro_overview}
\end{figure*}

Most EHR-based benchmarks consider single-document contexts \citep{Pampari2018emrQAAL, Yue2020CliniQG4QAGD}, yet in practice relevant information may span multiple documents. Reasoning over multiple long, temporally-dependent documents raises questions around Long-Context (LC) limitations and evidence positioning sensitivity \citep{liu2023lost, xiao2023efficient, Adams2024LongHealthAQ}. Moreover existing LC medical QA work is either synthetic, MCQA-based and short \citep{Adams2024LongHealthAQ} or small and not human-validated \citep{fan2024medodyssey}. 

We address this gap by investigating LLM capabilities on longitudinal, patient-centric EHRs with human-validated QA pairs.
While recent LLMs push the frontiers of context size, performance does not scale with increased context \citep{levy2024same, modarressi2025nolima}. RAG \citep{lewis2021rag} offers a cost-efficient alternative by selecting relevant context on demand, with reported gains over Full-Context (FC) across various settings \citep{zhang2024marathon, xu2024chatqa, Bai2024LongBenchVT}, though its advantage peaks at 32K for large LLMs \citep{leng2024long}. Models still struggle with specialized content such as EHRs \citep{fan2025medeureka, myers2025lessons, almannaa-etal-2026-optimed}, leaving Full-Context (FC) vs RAG superiority in LC clinical EHR-based QA an open question. Prior works comparing RAG and FC \citep{li-etal-2024-retrieval} or examining RAG's limits \citep{leng2024long} relied on a single hand-picked dense retriever, which may underestimate RAG's potential. EHR-specific retrieval work has shown the complementarity of dense and sparse retrievers in medical tasks \citep{fan2025medeureka, myers2025lessons} with hybrid retrieval outperforming both \citep{stuhlmann2025efficient}. We design a hybrid RAG pipeline composed of well-established robust components as a principled baseline enabling a fair comparison against FC prompting in LC settings. We make the following contributions:
\begin{itemize}[leftmargin=*]
    \setlength{\itemsep}{1pt}
    \setlength{\parskip}{2pt}
    \vspace{-0.5cm}
    \item We assess how LLMs of different capabilities and sizes perform on Long-Context clinical QA given longitudinal EHR patient notes, covering three task formulations: 1) MCQA, 2) Extractive, and 3) Open-ended generative QA (\S\ref{sec:datasets}, \S\ref{sec:ragsetup}) across a variety of LC and RAG settings (Fig.~\ref{fig:intro_overview}).
    \item We perform comprehensive experiments evaluating different note inclusion strategies, examining their effect on performance and model memorization (\S\ref{sec:contextform}).
    \item We evaluate RAG against Full-Context (FC) prompting over longitudinal EHR patient histories across task formulations and context sizes, demonstrating RAG superiority across tasks (\S\ref{sec:mainresults}, \S\ref{sec:quantllmaraganalysis}), through a designed hybrid RAG pipeline motivated by biomedical retrieval literature (\S\ref{sec:ragstrategies}), yielding insights into long-context EHR comprehension.
    \item Our multi-faceted evaluation methodology covers both single- and multi-document long-form reasoning settings, while a category-level quantitative analysis reveals where RAG and FC respectively excel and struggle, with RAG falling behind in longitudinal reasoning (\S\ref{sec:mainresults}, \S\ref{sec:quantllmaraganalysis}, 
\S\ref{sec:erroranalysis}).


\end{itemize}

\section{Related Work}

\noindent \textbf{Leveraging Long Context}: While increasing long-context window capabilities of LLMs, e.g. GPT-4o \citep{hurst2024gpt} (128K), Claude 3~\citep{CLAUDE3} (200K), Gemini 1.5 \citep{team2024gemini} (1 million), boost their Long-Context performance, they still struggle in general purpose open-form QA beyond short passages~\cite{zhang2024bench, chen2025longleader}. Despite positional embedding techniques like RoPE \citep{su2024roformer}, YaRN~\citep{peng2023yarn} and Position Interpolation \citep{chen2023extending}, QA task performance remains low even for 32K contexts~\citep{chen2025longleader}, with LLMs particularly challenged by long-range dependencies requiring reasoning~\citep{li2024loogle, dong2023bamboo, bai2023longbench, liu2023lost, hsieh2024ruler}. The performance gap between open-source and closed-source models~\citep{li2024loogle} further raises concerns for patient-centric settings where privacy matters. \citet{fan2024medodyssey} show notable performance drops in medical QA up to 200K tokens, yet despite the prominence of longitudinal context in real-world patient data, little work investigates long-form LLM capabilities.




\begin{figure*}
    \centering
    \includegraphics[width=0.8\linewidth]{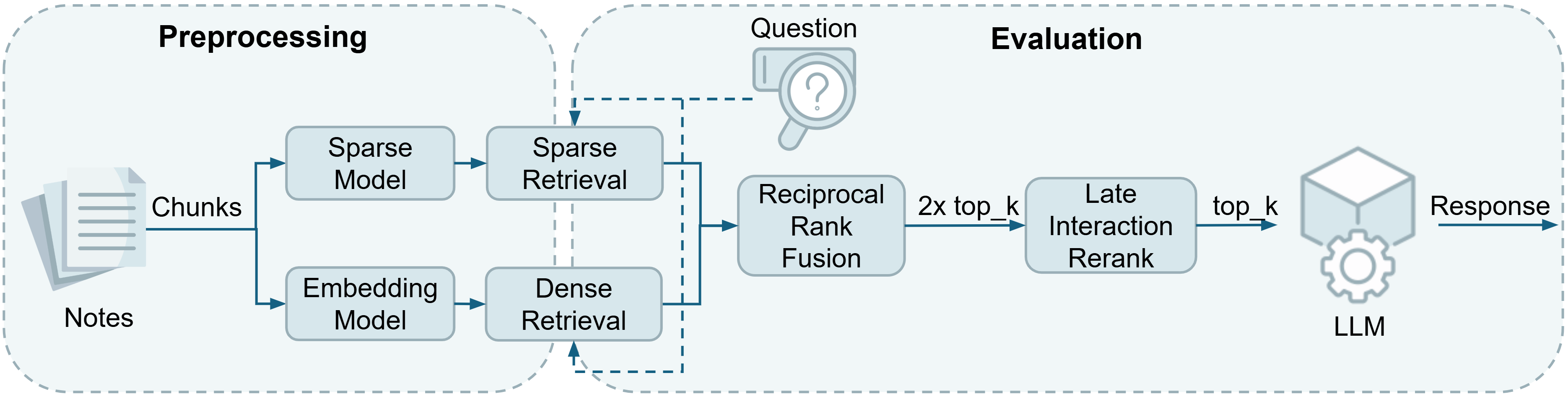}
    \caption{RAG Pipeline: clinical notes are split into 512-token chunks, encoded using dense (Qwen3-Embedding-8B) and sparse (BM25) embeddings, and indexed in Qdrant. At evaluation time the question retrieves 2$\times$top-$k$ chunks from each retriever, Reciprocal Rank Fusion (RRF) combines them into a total of 2$\times$top-$k$ chunks, and Late Interaction Reranking yields the final top-$k$ chunks, sorted chronologically before being passed to the LLM.}
    \label{fig:ragapproach}
\end{figure*}

\noindent \textbf{Retrieval-augmented Generation (RAG)} has been leveraged in a prompt-based plug-in manner to address black-box Long-Context challenges \citep{Yu2023ImprovingLM}. 
LLM-based retrievers like BGE M3 \citep{Chen2024BGEMM} and self-reflective approaches like Self-RAG \citep{Asai2023SelfRAGLT} have shown notable gains in open-domain QA. In the biomedical domain, MedCPT \citep{jin2023medcpt} and BMRETRIEVER \citep{xu2024bmretriever} achieve strong retrieval performance, while Med-RAG \citep{zhao2025medrag} and Self-BioRAG \citep{jeong2024improving} further improve EHR-based diagnostics through knowledge graph reasoning and self-reflection respectively. Yet despite strong in-domain retrieval capabilities, models still struggle with specialized medical content such as EHRs \citep{fan2025medeureka, myers2025lessons}, motivating a thorough examination of RAG versus FC over realistic long-context patient-oriented QA. Instead of relying on a single hand-picked  retriever, we design a literature-grounded principled hybrid RAG pipeline \citep{ryan2026self, stuhlmann2025efficient, sohn2025rationale} to enable a fair and robust comparison against FC on longitudinal EHR-based QA (see \S \ref{sec:ragstrategies}).

\section{Approach}

Since strong medical QA performance offers the potential of LLM integration in clinical workflows, we aim to uncover LLM limitations given complex patient-oriented scenarios. We focus on \textbf{content-comprehension} settings using longitudinal EHRs, 
thus emphasizing relevant patient-centric \textbf{long-context} across time and documents.

Our comprehensive approach spans across
expert annotated datasets covering \textbf{three task formulations}: 1) MCQA, 2) Extractive, and 3) Open-ended QA. We assess each dataset across \textbf{four settings 
with varying note relevance} (\S\ref{sec:contextform}).

\subsection{Context Inclusion Formulation} 
\label{sec:contextform}

Most datasets use a single note/report or up to a few discharge summaries, leaving a large amount of \textit{longitudinal patient supplementary} content unused in the EHR history. We aim to assess the significance of the patient's longitudinal Full-Context (FC), evaluating a realistic clinical setting with no a-priori knowledge of the most relevant note(s). Therefore we leverage the MIMIC resources to augment the context with each patient's longitudinal notes\footnote{MIMIC preprocessing is described in Appendix \ref{sec:data-prep}.}, resulting in Long Context (LC) from the entire patient's note history (\textit{Include All}). We then investigate how LLMs leverage long-form context to answer patient-specific questions. 

We explore four data inclusion scenarios depicted in Fig. \ref{fig:intro_overview}, allowing comprehensive analysis: 
\begin{enumerate}[leftmargin=*]
    \setlength{\itemsep}{1pt}
    \setlength{\parskip}{1pt}
    \item \textbf{Exclude All}: Exclusion of all notes - assessing model note memorization.
    \item \textbf{Exclude Related}: Exclusion of the identified most relevant notes - assessing the informational value of only the supplementary patient notes for QA, without including the note(s) identified as most relevant, thus isolating the contribution of longitudinal supplementary context.
    \item \textbf{Include All}: Inclusion of all patient notes - assessing LC processing 
    of varying importance. 
    \item \textbf{Include Related}: Inclusion of only the most relevant note(s) (DS, CN or RR) for the question from the patient's MIMIC context as defined by each dataset's release (see \textit{Answer Location} in Table \ref{tab:qa_datasets}) - assessing the model's ability to leverage the most relevant context.
\end{enumerate}
These settings enable direct comparison between memorization,  supplementary context utility and full versus targeted context 
inclusion, spanning from no context to the full longitudinal 
history.

\subsection{Dataset Selection Criteria} \label{sec:datasetselect}

The following criteria were used for benchmark selection (full comparison in Table \ref{tab:dataset_landscape}):

\noindent \textbf{Beyond MCQA}:
Medical MCQA benchmarks i.e. PubMedQA \citep{jin2019pubmedqa}, MedQA \citep{jin2021disease}, MedMCQA \citep{pal2022medmcqa} 
evaluate option selection instead of grounded synthesis and risk pretraining leakage from textbooks, which can inflate scores, providing little signal towards noisy clinical settings. We go beyond MCQA evaluating on both extractive and open-ended QA.

\noindent \textbf{Long-Context Documents}: 
To address LLM capabilities in handling real-world LC we seek benchmarks extending well-beyond 8K, with inclusion of patient-specific content.

\noindent \textbf{Expert Annotated QA pairs} 
grounded on multi-document EHRs were our choice, to avoid clinical reliability issues like non-expert curation \citep{fan2024medodyssey} or synthetic data \citep{Adams2024LongHealthAQ}.

\noindent \textbf{Patient-centric Longitudinal Content}:
Prioritizing clinically-relevant QA with reasoning, we focus on MIMIC-based EHR datasets \citep{Johnson2016,Johnson2023}, offering timestamped notes across document types with consistent format. 

\subsection{RAG Approach} \label{sec:ragstrategies}

Retrievers are distinguished into \textit{sparse} and \textit{dense}. Sparse retrievers such as BM25 \citep{robertson2009probabilistic} and Splade \citep{formal2021splade} use overlapping terms to match queries with snippets, while dense retrievers perform matches via semantic similarity in the embedding space. Using LLM-based embeddings 
for dense retrieval has become standard practice. 

Research on biomedical MCQA \citep{wang2024augmenting} and biomedical document retrieval \citep{luo2022improving} has shown that hybrid approaches combining sparse and dense retrieval outperform single component counterparts. Hybrid sparse-dense retrieval with Reciprocal Rank Fusion (RRF) has further demonstrated strong performance across domains \citep{arivazhagan2023hybrid, lee2023complementarity}, while recent work across biomedical QA shows that hybrid retrieval outperforms both sparse and dense retrievers alone \citep{stuhlmann2025efficient}, with hybrid retrieval with RRF specifically demonstrating superiority on PubMedQA and MedQA \cite{ryan2026self}. 

Additionally, related EHR work has further shown that no single embedding model dominates across medical retrieval tasks \citep{myers2025lessons}, with dense and sparse retrievers offering complementary benefits depending on the setting and dataset \citep{fan2025medeureka}. Finally, re-ranking of retrieved snippets also exhibits superior performance in biomedical applications \citep{jin2023medcpt,wang2024augmenting,sohn2025rationale}. FC versus RAG comparisons \citep{li-etal-2024-retrieval, leng2024long} rely on a single dense retriever, potentially underestimating RAG's potential. Provided substantial evidence for the superiority of hybrid retrieval over single-component counterparts in this domain, we leverage a hybrid approach followed by re-ranking, presented in Figure \ref{fig:ragapproach}, focusing on our primary goal on evaluating LLM capabilities under different context settings on long longitudinal EHR patient data. 

\noindent Clinical notes are segmented into 512-token chunks, forming the document collections $D = \{d_1, d_2, \dotsc , d_n\}$. The question respectively forms the query $Q$. 

\noindent\textit{Sparse retrieval}: we employ a lexical retriever as a ranking function of each chunk $d_j$ 
\footnote{Here terms document and chunk are used interchangeably.}:
 \begin{equation}
    \small S_{\mathrm{sparse}}(Q,d_j)\;=\;\sum_{i=1}^{n} \mathrm{IDF}(q_i)\cdot \tfrac{f_{q_i,d_j}\,(k_1+1)}{f_{q_i,d_j}+k_1\!\left(1-b + b\cdot \frac{|d_j|}{\mathrm{avgdl}}\right)}
    \label{eq:bm25}
\end{equation}

\noindent \textit{Dense retrieval}: we use the same encoder, $E$, for the query and the chunks and capture semantic similarity of such pairs using the cosine similarity:
    \begin{equation}
    S_{\mathrm{dense}}(Q,d_j) \;=\; \frac{E(Q)\cdot E(d_j)}{\lVert E(Q)\rVert\,\lVert E(d_j)\rVert}
    \label{eq:dense}
    \end{equation}
   
\noindent \textit{Reciprocal Rank Fusion (RRF)}: combines the top-$k$ chunks from each retriever, through $\mathrm{rank}_s(d_j)$ denoting the rank assigned to document $d_j$ by retriever $s \in \{\text{dense}, \text{sparse}\}$, using a smoothing constant $k_{RRF}$:
    \begin{equation}
    \mathrm{RRF}(d_j) \;=\; \sum_{s\in \mathcal{S}} \frac{1}{k_{RRF} + \mathrm{rank}_s(d_j)}
    \label{eq:rrf}
    \end{equation}

\noindent \textit{Late Interaction Reranking}: measures how well every $Q$-token, $\mathbf{q}_i$, is supported by at least one document token, $\mathbf{d}_{j,r}$, to produce the top-$k$ set:
    \begin{equation}
    S_{\mathrm{MaxSim}}(Q,d_j) \;=\; \sum_{i=1}^{|Q|}\; \max_{1\le r \le |d_j|} \langle \mathbf{q}_i, \mathbf{d}_{j,r} \rangle
    \label{eq:colbert}
    \end{equation}
    
\noindent \textit{Document Ordering}: we 
sort retrieved chunks chronologically by note timestamp rather than retrieval score, as temporal ordering yields superior long-context performance \citep{yu2024defense}. 

\definecolor{lightgray}{rgb}{0.9, 0.9, 0.9}
\definecolor{lightergray}{rgb}{0.8, 0.8, 0.8}

\begin{table*}
\begin{adjustbox}{max width=1\textwidth,center}
  \centering
  \begin{tabular}{cccccccccc}
    \toprule
    \rowcolor{lightergray}
\textbf{Dataset} & \textbf{Context Data} & \textbf{QA Pairs} & \textbf{Patients} & \textbf{Mean/Max Context} & \textbf{Task Types} & \textbf{Answer Location}  & \textbf{Reasoning}  & \textbf{Question Generation} & \textbf{Answer Annotation}\\
    \midrule
    \midrule
   \Large CliniQG4QA  & \multirow{2}{*}{MIMIC-III}  & 1{,}287& 36   & 4K / 7K & \multirow{2}{*}{Extractive}&  Clinical Note   & $\Large\color{gray}{\pmb\times}$ & Experts & Clinical experts (3) \\
   \Large RadQA   &   & 3{,}509& 80   & 6K / 14K   & & Radiology Report  & Single-Note&  Human-generated & Physicians (2) \\
    \midrule
    \Large EHR-DS-QA   & \multirow{2}{*}{MIMIC-IV}  & 478    & 70   & 46K / 131K & Open-ended & Clinical Note/Dis. Summary & $\Large\color{gray}{\pmb\times}$ &LLMs & Physician-verified (1) \\
    \Large EHRNoteQA   &   & 962    & 962  & 9K / 39K   & MC,Open-ended  & Dis. Summaries & Multi-Note& LLM & Clinician-refined (3) \\   
    \bottomrule
  \end{tabular}
  \end{adjustbox}
  \caption{EHR QA Dataset Statistics. Mean/Max Context corresponds to the \textit{Include All} setting.}
  \label{tab:qa_datasets}
\end{table*}

\subsection{Evaluation Methodology} \label{sec:evalframework} 

Evaluating medical QA systems has traditionally relied on single metrics and MCQA benchmarks \cite{jin2019pubmedqa, jin2021disease, pal2022medmcqa}, which probe memorized knowledge and underrepresent the complexity 
of clinical reasoning, 
and EHR-grounded context. Aiming for a clinically meaningful assessment, we adopt a multi-dimensional, reference-based scheme that complements lexical overlap with embedding-based semantic similarity and domain-adapted Natural Language Inference (NLI). 
In line with emerging clinical evaluation practices (MEDIC \citep{kanithi2024medic}) and frameworks (ArchEHR \citep{soni2025archehr}) and the rise of LLM-as-a-judge, we employ a calibrated rubric 
and we prioritize widely available metrics while cross-checking signals for robustness.


Our datasets comprise gold-standard QA pairs that are either expert-verified or expert-generated, thus enabling a clinically centric evaluation:
\vspace{.2em}

\noindent\textbf{METEOR} \cite{banerjee2005} to capture \textit{surface-level} and \textit{lexical similarity}.
\vspace{.2em}

\noindent\textbf{BERTScore} \cite{bert-score}, computed with Clinical BioBERT embeddings \cite{alsentzer2019publicly} to assess \textit{semantic similarity}. 
\vspace{.2em}

\noindent\textbf{NLI Scores}, to capture the logical relationship between reference and candidate answers. We use a domain-adapted NLI model \cite{deka2023multiple} 
and then measure \textit{Factual Consistency} using the probability of non-contradiction \cite{song2024combining} and \textit{Factual Precision} using the entailment probability.

\vspace{.2em}

\noindent\textbf{LLM-as-a-judge}, employed with a 5-point scale rubric on three aspects: 
\textit{Correctness} capturing factual consistency with respect to the gold answer, penalizing contradictions while allowing compatible additional information;~\textit{Completeness} capturing recall by assessing the predicted answer’s coverage of information present in the reference; \textit{Faithfulness} capturing precision assessing if the predicted answer only contains information that is supported by the reference. Prompts are provided in Appendix~\ref{sec:judge_prompts}.


\vspace{.2em}
\noindent \textbf{Accuracy} reported on the MCQA task.

Our metric selection is grounded in correlation analysis (Figures \ref{fig:correlation_matrix}, \ref{fig:correlation_matrix_long} in Appendix \ref{sec:appendix.metric_correlation}), prioritizing complementary metrics with lower inter-correlation to maximize signal diversity and minimize redundancy.
Metrics correlate stronger within families, while  METEOR and LLM Faithfulness are also highly correlated with BERTScore. In the long-context high-correctness regime, correlations are weaker. Thus, in the main tables we  report one metric per family for a comprehensive yet non-redundant view across complementary dimensions.


\section{Experiments}




\subsection{Datasets}
\label{sec:datasets}

We leverage MIMIC-III \citep{Johnson2016} and MIMIC-IV \citep{Johnson2023}, comprising de-identified EHRs covering hospital admissions and ICU stays, as the underlying context. They support single- and multi-note settings 
and include: clinical notes (CN), discharge summaries (DS) and radiology reports (RR) per patient over time. 



Our four selected QA datasets only use QA pairs curated or validated by clinical experts to ensure medical validity. We provide a brief dataset overview below, also summarized in Table~\ref{tab:qa_datasets}:  

\noindent \textbf{CliniQG4QA} \citep{Yue2020CliniQG4QAGD}: An extractive span benchmark on MIMIC-III comprising 1,287 QA pairs generated or verified by three clinical experts. We only use the expert-annotated set.

\noindent \textbf{RadQA} \citep{Soni2022}: An extractive dataset of 3,074 physician-crafted questions and 6,148 answer spans from Findings and Impressions sections of radiology reports in MIMIC-III, annotated by two human annotators. After filtering unanswerable pairs we retain 3,509 pairs.

\noindent \textbf{EHR-DS-QA} \citep{EHRDSQA2023}: A synthetic QA corpus generated by two LLMs on discharge summaries and clinical notes from MIMIC-IV. Of 506 physician-verified pairs, we retain 478 marked as correct.

\noindent \textbf{EHRNoteQA} \citep{Kweon2024EHRNoteQAAL}: Built on MIMIC-IV, comprising 962 QA pairs authored by GPT-4 and iteratively refined by three clinicians, each corresponding to multiple discharge summaries ($\sim$2.3 per patient) and supporting both open-ended and MCQA formats.



\begin{table*}[t]
\centering
\begin{adjustbox}{max width=\textwidth,center}
\begin{tabular}{@{}lccccccccccccccc@{}}
\toprule
\textbf{Model} & \multicolumn{1}{c}{\textbf{Setting}} & \multicolumn{3}{c}{\textbf{EHRNoteQA}} & & \multicolumn{3}{c}{\textbf{EHR-DS-QA}} & \multicolumn{3}{c}{\textbf{RadQA}} & \multicolumn{3}{c}{\textbf{CliniQG4QA}} \\
 & & \multicolumn{3}{c}{Open-ended} & MC & \multicolumn{3}{c}{Open-ended} & \multicolumn{3}{c}{Extractive} & \multicolumn{3}{c}{Extractive} \\
 & & LLM & NLI & F1 & Acc. & LLM & NLI & F1 & LLM & NLI & F1 & LLM & NLI & F1 \\
\midrule
HuatuoGPT-o1 7B & \multirow{4}{*}{Exclude All}     & \underline{7.85} & \textbf{23.36} & \textbf{68.82} & 51.08 & \textbf{26.65} & \textbf{33.86} & 72.21 & \underline{25.06} & \underline{27.90} & \underline{63.78} & \underline{15.96} & \underline{23.01} & \underline{64.44} \\
Qwen2.5 7B      & & \textbf{9.73} & \underline{18.18} & \underline{68.41} & 51.46 & \underline{24.79} & \underline{33.00} & \underline{72.35} & \textbf{39.08} & \textbf{29.24} & \textbf{70.73} & \textbf{27.02} & \textbf{26.30} & \textbf{74.06} \\
Qwen2.5 32B     & & 0.71 & 11.30 & 66.56 & \underline{58.34} & 20.16 & 20.23 & \textbf{72.67} & 3.86 & 2.88 & 10.35 & 0.11 & 0.19 & 2.16 \\
QwQ 32B         & & 2.68 & 12.23 & 66.98 & \textbf{59.24} & 24.16 & 23.80 & 71.70 & 1.51 & 6.44 & 15.29 & 0.93 & 5.62 & 14.77 \\
\midrule
HuatuoGPT-o1 7B & \multirow{4}{*}{Exclude Related}  & \underline{26.62} & \textbf{40.26} & 72.57 & 58.35 & \underline{28.49} & \textbf{33.98} & 71.84 & \underline{24.39} & \underline{28.28} & \underline{63.33} & \underline{15.70} & \underline{24.13} & \underline{64.15} \\
Qwen2.5 7B      & & 25.24 & \underline{28.64} & 72.72 & \underline{59.86} & \textbf{29.90} & \underline{28.17} & \underline{73.35} & \textbf{38.86} & \textbf{29.07} & \textbf{70.72} & \textbf{27.01} & \textbf{25.95} & \textbf{74.09} \\
Qwen2.5 32B     & & 25.59 & 23.67 & \underline{73.01} & 58.72 & 22.16 & 22.35 & 73.15 & 3.77 & 2.78 & 10.02 & 0.13 & 0.20 & 2.17 \\
QwQ 32B         & & \textbf{27.46} & 23.01 & \textbf{74.35} & \textbf{60.15} & 27.19 & 23.98 & \textbf{73.51} & 1.14 & 7.11 & 17.48 & 0.99 & 4.84 & 15.75 \\
\midrule
HuatuoGPT-o1 7B & \multirow{4}{*}{Include All}     & \underline{72.60} & 45.45 & 78.21 & 78.94 & \underline{59.76} & \textbf{63.66} & 77.14 & 63.92 & 49.41 & 76.38 & 67.85 & 52.45 & 80.08 \\
Qwen2.5 7B      & & 70.49 & 38.11 & 79.95 & 78.81 & \textbf{62.49} & 61.00 & \underline{79.14} & 62.56 & \underline{50.13} & 76.65 & 68.89 & 54.13 & 81.27 \\
Qwen2.5 32B     & & 67.88 & \textbf{55.33} & \textbf{81.50} & \cellcolor{orange!25}\textbf{90.97} & 57.21 & \underline{61.96} & \textbf{79.59} & \textbf{67.74} & 50.10 & \textbf{77.55} & \cellcolor{orange!25}\textbf{80.49} & \underline{59.22} & \underline{83.80} \\
QwQ 32B         & & \textbf{75.52} & \underline{47.14} & \underline{81.02} & \underline{89.25} & 57.68 & 60.54 & 78.90 & \underline{67.46} & \textbf{53.26} & \underline{77.47} & \underline{78.95} & \cellcolor{orange!25}\textbf{66.00} & \textbf{83.94} \\
\midrule
HuatuoGPT-o1 7B & \multirow{4}{*}{Include Related}  & 70.73 & 39.30 & 79.71 & 76.87 & \underline{63.46} & 60.65 & 77.20 & 64.37 & 49.19 & 76.48 & 70.47 & 53.19 & 79.83 \\
Qwen2.5 7B      & & 74.44 & 39.15 & 80.78 & 80.9 & \cellcolor{orange!25}\textbf{64.76} & \cellcolor{orange!25}\textbf{65.14} & \underline{80.28} & 63.01 & 50.25 & 76.71 & 69.06 & 54.87 & 81.29 \\
Qwen2.5 32B     & & \underline{76.01} & \cellcolor{orange!25}\textbf{61.41} & \cellcolor{orange!25}\textbf{82.48} & \textbf{90.33} & 59.12 & \underline{64.86} & \cellcolor{orange!25}\textbf{80.70} & \underline{67.76} & \underline{50.78} & \underline{77.67} & \textbf{79.81} & \underline{59.21} & \underline{83.55} \\
QwQ 32B         & & \cellcolor{orange!25}\textbf{82.03} & \underline{47.19} & \underline{82.36} & \underline{87.57} & 61.31 & 61.68 & 79.70 & \cellcolor{orange!25}\textbf{67.79} & \cellcolor{orange!25}\textbf{55.48} & \cellcolor{orange!25}\textbf{77.69} & \underline{79.74} & \textbf{65.86} & \cellcolor{orange!25}\textbf{84.35} \\
\bottomrule
\end{tabular}
\end{adjustbox}
\captionsetup{justification=centering}
\caption{Results for Full Context in each setting across datasets. 
\textbf{Bold} is best and~\underline{underlined} the second best model in each Setting and Metric. Orange highlights the global best across all Settings per Metric. \textit{LLM} corresponds to LLM Correctness, \textit{NLI} to NLI Entailment and \textit{F1} to BERTScore (BioBERT) F1.}
\label{tab:qa_results}
\end{table*}

\noindent \textbf{Context Bins}: For \textit{Include All} we report performance across four token bins: 0--8K, 8--16K, 16--32K, and 32--128K. As shown in Appendix~\ref{app:contextlen}, both \textit{Include Related} and \textit{Exclude Related} yield shorter contexts across all bins, most notably in the 16--32K bin for \textit{Include Related}.

\subsection{Models and Experimental Setup} \label{sec:ragsetup}

\noindent \textbf{LLMs}: To explore the effect of model size, domain specialization and reasoning capabilities we study four open Qwen2.5~\citep{team2024qwen2} single-family models, ensuring a controlled comparison: 
\begin{itemize}[leftmargin=*]
\setlength{\itemsep}{1pt}
\setlength{\parskip}{3pt}  
\item \textit{Qwen2.5-7B-Instruct}
\item \textit{HuatuoGPT-o1-Qwen-7B} \citep{chen2024huatuogpt}: medical LLM based on Qwen2.5-7B-Instruct
\item \textit{Qwen2.5-32B-Instruct-128K} 
\item \textit{QwQ:32B}: reasoning model
\end{itemize}

\noindent \textbf{Evaluation}: Through comparisons against Selene-8B \citep{alexandru2025atla} and Prometheus-8x7B v2.0 \citep{kim2024prometheus}, we finally  selected \textsc{Qwen-2.5-32B-Instruct} for the LLM-as-a-judge for stability and agreement (see Appendix \ref{sec:judgecomps}). 

\vspace{0.5em}

\noindent \textbf{Retrieval Settings}: 
Based on biomedical QA retrieval research that demonstrates the superiority of combining sparse and dense retrievers (see \S\ref{sec:ragstrategies}), we designed a hybrid retrieval approach to handle the semantic, lexical, and query complexity. The selected model components are listed below:

\begin{itemize}[leftmargin=*]
\setlength{\itemsep}{1pt}
\setlength{\parskip}{3pt}  
\item \textit{Dense Retrieval} using the open embedding model Qwen3-Embedding-8B \citep{zhang2025qwen3}.

\item \textit{Sparse Retrieval} using BM25, selected after ablating two sparse retrievers (see Table \ref{tab:retrievers_comp}).

\item \textit{Late Interaction Reranking} with Reason-ModernColBERT, a ColBERT \citep{khattab2020colbert} model finetuned on the ReasonIR dataset \citep{shao2025reasonir} demonstrating high performance on a reasoning-intensive retrieval benchmark \citep{su2025bright}. 
\end{itemize}

\noindent We explore two chunk inclusion strategies: i) direct chunk inclusion (\textit{RAG}), ii) hierarchical whole parent note inclusion of the retrieved chunk for full-note context view (\textit{RAG HIR}). Under both settings, the retrieved content is sorted chronologically using the corresponding MIMIC note creation timestamps before being concatenated into the prompt (\S\ref{sec:ragstrategies}). Therefore the model receives a patient timeline. Retrieval is based on 512-token chunks selected through chunking strategy experimentation (Appendix~\ref{sec:chunks}, Table~\ref{tab:chunks}).

\vspace{0.5em}

\noindent \textbf{Experimental Setup}: LLM inference and prompt formulation specifics are in Appendices~\ref{sec:compute} and~\ref{sec:prompt_specs}.



\section{Results and Discussion}
\label{sec:results}

\subsection{Main Results}
\label{sec:mainresults}

Performance for the different inclusion settings across
datasets and models is shown in Table \ref{tab:qa_results}. Statistical significance of the key claims drawn from these results is examined through Philipp Koehn's paired bootstrap resampling tests (Appendix \ref{sec:sigtests}).








\paragraph{Model Size, Tasks and Inclusion Settings:}

\begin{figure}[htbp]
\centering
\includegraphics[width=\columnwidth]{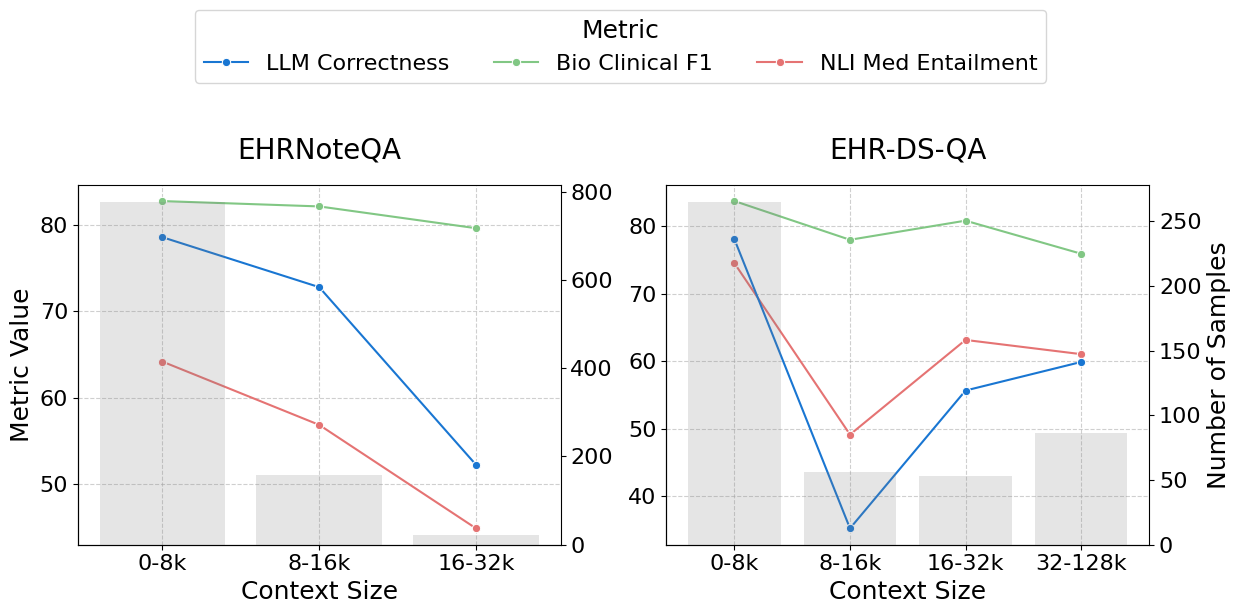}
\captionsetup{justification=centering}
\caption{Metric performance over context size on \textit{Include All} for Qwen2.5:32B-Instruct.}
\label{fig:metric_performance_context}
\end{figure}

Consistent with general findings, larger models perform better across all tasks (Table~\ref{tab:qa_results}), with the performance gap particularly pronounced in tasks requiring reasoning or information synthesis from multiple sources, namely EHRNoteQA and RadQA. However, smaller models perform better in settings excluding relevant information or completely removing the context, suggesting stronger memorization as also strongly suggested by Philipp Koehn’s paired bootstrap resampling significance test \citep{koehn2004statistical} reported in Table \ref{tab:sig_memorization} (performed on open-ended generative datasets). Unlike the rest of the tasks, Multiple-Choice (MC) shows strong evidence of memorization with a performance of 59.24\% in the \textit{Exclude All} setting, while MC demonstrates 90\%+ performance, open-ended and extractive tasks remain challenging for LLMs.

\begin{table}
\begin{adjustbox}{max width=1\linewidth,center}
\begin{tabular}{@{}llcccccccc@{}}
\toprule
\textbf{Model} & \textbf{Setting} & \multicolumn{4}{c}{\textbf{EHRNoteQA}} & \multicolumn{3}{c}{\textbf{EHR-DS-QA}} \\
 &  & \multicolumn{3}{c}{Open-ended} & MC & \multicolumn{3}{c}{Open-ended} \\
 & & LLM & NLI & F1 & Acc. & LLM & NLI & F1 \\
\midrule
\multirow{7}{*}{HuatuoGPT-o1 7B} & Include All & 70.68 & 39.52 & 77.26 & 75.36 & 50.25 & 59.01 & 75.30 \\
&Include Related &66.88 & 29.62 & \underline{79.37} & 72.38 & 56.51 & 54.71 & 75.29 \\
\addlinespace
 & RAG 5       & 59.94 & \cellcolor{orange!25}\textbf{61.34} & 78.44 & \underline{75.67} & 57.57 & 61.53 & \underline{78.22} \\
 & RAG 10      & 61.03 & 48.51 & 77.99 & 73.4 & \underline{58.08} & \underline{62.45} & \textbf{78.61} \\
 & RAG 15      & 60.84 & 52.53 & 77.92 & 74.97 & \cellcolor{orange!25}\textbf{61.95} & \cellcolor{orange!25}\textbf{64.84} & 77.68 \\
 \addlinespace
 & RAG HIR 3   & 70.13 & 56.91 & 77.99 & 70.81 & 50.06 & 53.82 & 76.17 \\
 & RAG HIR 5   & \textbf{70.92} & \underline{57.51} & \textbf{79.52} & 72.38 & 48.46 & 55.21 & 76.66 \\
 & RAG HIR 7   & \underline{70.80} & 54.61 & 79.35 & \textbf{80.46} & 55.99 & 57.12 & 77.35 \\
\midrule
\multirow{7}{*}{Qwen2.5 7B} & Include All & 66.86 & 28.36 & 79.23 & 75.04 & 54.26 & 52.78 & 77.07 \\
&Include Related &\underline{72.37} & 29.50 & 80.34 & 78.82 & \underline{58.60} & \underline{60.79} & 78.85 \\
\addlinespace
 & RAG 5       & 59.54 & 40.33 & 78.82 & 75.67 & 54.17 & 56.11 & \textbf{79.82} \\
 & RAG 10      & 64.03 & 45.02 & 79.05 & 73.40 & 57.11 & \textbf{62.76} & 79.60 \\
 & RAG 15      & 58.07 & \underline{52.08} & 79.20 & 75.60 & \textbf{59.25} & 60.32 & \underline{79.65} \\
 \addlinespace
 & RAG HIR 3   & 65.54 & 46.34 & \underline{80.37} & 76.93 & 50.73 & 53.12 & 78.65 \\
 & RAG HIR 5   & \textbf{72.68} & \textbf{54.69} & 80.17 & \underline{79.76} & 53.59 & 52.56 & 77.66 \\
 & RAG HIR 7   & 67.55 & 46.41 & \textbf{80.69} & \textbf{84.31} & 52.10 & 52.39 & 78.26 \\
\midrule
\multirow{7}{*}{Qwen2.5 32B}  & Include All & 62.54 & 50.90 & 80.87 & 89.79 & 50.27 & 57.78 & 78.22 \\
&Include Related & \cellcolor{orange!25}\textbf{74.16} & \textbf{59.11} & \underline{82.17} & 88.78 & 51.88 & 59.78 & 79.48\\
\addlinespace
 & RAG 5       & 60.35 & 46.36 & 80.28 & 80.46 & \underline{58.02} & 58.91 & 80.29 \\
 & RAG 10      & 67.00 & 54.54 & 80.47 & 77.87 & \textbf{60.52} & 60.31 & 80.24 \\
 & RAG 15      & \underline{73.45} & 53.94 & 80.84 & 81.02 & 55.68 & \underline{61.39} & \cellcolor{orange!25}\textbf{80.71} \\
 \addlinespace
 & RAG HIR 3   & 67.48 & \underline{57.17} & \cellcolor{orange!25}\textbf{82.41} & 83.36 & 55.13 & \textbf{63.32} & \underline{80.55} \\
 & RAG HIR 5   & 68.86 & 54.87 & 81.59 & \underline{90.74} & 51.20 & 60.73 & 80.12 \\
 & RAG HIR 7   & 72.00 & 47.33 & 81.84 & \cellcolor{orange!25}\textbf{91.68} & 52.82 & 57.27 & 80.21 \\
\bottomrule
\end{tabular}
\end{adjustbox}
\captionsetup{justification=centering}
\caption{Open-ended and MCQA Results for Context of 8K+ tokens including LC and RAG Methods. \textbf{Bold} is best and  \underline{underlined} the second best performance per Model and Metric. Orange highlights the global best across all models per Metric.}
\label{tab:qa_results_rag}
\end{table}

The \textit{Exclude All} setting primarily assesses model memorization, with Qwen2.5-7B and HuatuoGPT-o1-7B outperforming Qwen2.5-32B further indicating stronger memorization in smaller models. The \textit{Exclude Related} setting, which includes only supplementary patient history excluding the most relevant notes, yields near-identical performance to \textit{Exclude All} across all models and sizes for the purely extractive RadQA and CliniQG4QA datasets, where the answer is contained within the single related note, rendering supplementary context uninformative. By contrast, for the open-ended generative datasets, \textit{Exclude Related} outperforms \textit{Exclude All} by a margin, demonstrating the usefulness of supplementary patient material as also strongly supported by
the significance tests in Table~\ref{tab:sig_supplementary}. The \textit{Include Related} setting outperforms \textit{Include All} on most datasets and metrics except for MC, while the gap between the two suggests that LLMs struggle to identify the most relevant information in the full \textit{Include All} context.




\paragraph{Reasoning and Fine-tuning:}
\emph{The QwQ:32B general reasoning model} showed improved results on Table \ref{tab:qa_results} over other models overall on RadQA, EHRNoteQA (the reasoning datasets), suggesting that reasoning models can offer an advantage in tasks that demand complex inference across multiple sources of information, particularly at larger model sizes where reasoning capabilities can be effectively leveraged. 
Significance testing (Table~\ref{tab:sig_reasoning}) confirms that QwQ-32B significantly outperforms Qwen2.5-32B on LLM Correctness for multi-note QA (EHRNoteQA) but not on NLI or BioClinical F1, nor on any metric for single-note QA (EHR-DS-QA). This suggests reasoning models generate answers that go beyond the gold reference while remaining factually compatible, penalized by the reference alignment of NLI and BioClinical F1 (\S\ref{sec:erroranalysis}).

By contrast, \emph{the medically fine-tuned reasoning model, HuatuoGPT-o1 7B}, is statistically indistinguishable from the instruction-tuned Qwen2.5:7B-Instruct across metrics and open-ended datasets (Table~\ref{tab:sig_finetune}). We thus report an \emph{absence of measurable benefit} from medical fine-tuning, consistent with reported reduced performance of biomedical LLMs \citep{dorfner2024biomedical,dada2025does}. 


\begin{table}
\begin{adjustbox}{max width=1\linewidth,center}
\begin{tabular}{@{}llcccccccc@{}}
\toprule
\textbf{Model} & \textbf{Setting} & \multicolumn{3}{c}{\textbf{RadQA}} & \multicolumn{3}{c}{\textbf{CliniQG4QA}} \\
 & & LLM & NLI & F1 & LLM & NLI & F1 \\
\midrule
\multirow{2}{*}{HuatuoGPT-o1 7B} & Include All & 35.60 & \cellcolor{orange!25}\textbf{48.81} & 75.61 & 43.73 & 50.44 & 78.97 \\
 & RAG 5  & \textbf{36.61} & 29.51 & \textbf{76.61} & \textbf{44.20} & \textbf{54.51} & \textbf{79.54} \\
\midrule
\multirow{2}{*}{Qwen2.5 7B} & Include All & 36.10 &  \textbf{48.59} & 75.92 & 44.55 & 53.26 & 80.04 \\
 & RAG 5  & \cellcolor{orange!25}\textbf{37.09} & 28.39 & \cellcolor{orange!25}\textbf{77.10} & \cellcolor{orange!25}\textbf{47.03} & \cellcolor{orange!25}\textbf{58.36} & \cellcolor{orange!25}\textbf{81.73} \\
\bottomrule
\end{tabular}
\end{adjustbox}
\captionsetup{justification=centering}
\caption{Extractive QA Results for Context of 4K+ tokens. \textbf{Bold} is best performing per Model and Metric. Orange highlights the global best per Metric.}
\label{tab:qa_results_rag_extractive}
\end{table}

\paragraph{Context Size:}
Fig.~\ref{fig:metric_performance_context} shows that performance decreases with increased context size on open-ended QA, particularly for multi-note reasoning (EHRNoteQA), with similar trends across models (see Appendix, Fig.~\ref{fig:model_performance_context}). Notably, EHR-DS-QA exhibits a performance dip at 8-16K tokens across metrics and models, attributable to data noise and some ambiguous gold answers, consistent with observations by \citet{fan2024medodyssey} and \citet{ma2025beyond}.




\paragraph{RAG Performance:} Table~\ref{tab:qa_results_rag} shows two RAG settings: \textit{RAG $x$} and \textit{RAG HIR $x$}, with $x$ corresponding to the number of child 512-token chunks retrieved\footnote{We excluded QwQ:32B due to time and budget constraints}. RAG
demonstrates clear improvement on the single-note EHR-DS-QA task, where direct chunk inclusion consistently performs best. \textit{RAG HIR} is superior for the more complex multi-note reasoning of EHRNoteQA including MC for the smaller models, while larger models
benefit less from retrieval in complex reasoning settings. Significance testing in Table~\ref{tab:sig_rag} performed on the best on average RAG strategy per dataset and model confirms these findings. Impressively, RAG surpasses \textit{Include Related} performance across most settings, suggesting that effective retrieval can match or exceed having the most relevant note available a priori, with the exception of the multi-note reasoning dataset and larger models. Findings are consistent across 8K+ context sizes (see Fig.~\ref{fig:rag_performance_context}), with smaller models benefiting most. The medically fine-tuned reasoning model HuatuoGPT-o1 7B scores the best results on multiple metrics across the open-ended datasets, showing that RAG could benefit even more the smaller models that struggle in Full-context (FC) settings \textit{Include All}. In Table~  \ref{tab:qa_results_rag_extractive} we only evaluate 7B sized models using a single RAG setting, due to the task nature and limited context size of the extractive datasets. While the extractive datasets have shorter contexts overall, RAG still yields the best performance.


\subsection{Quantitative Analysis: RAG vs FC}
\label{sec:quantllmaraganalysis}

We first isolate challenging cases of metric disagreement (LLM Correctness, NLI Entailment, and BioClinical F1) across the open-ended QA EHRNoteQA and EHR-DS-QA. Metrics were first z-score standardized and samples were gathered for cases where some metrics belonged to the top 60th percentile while another metric fell within the bottom 40th percentile. The resulting 41 samples were manually categorized across QA types to determine which answer between RAG and FC is more reliable. We form the following rubric based on the manual analysis and show results in Table \ref{tab:rag_vs_fc_disagreement}:


\begin{itemize}[nosep,leftmargin=*]
    \item \textbf{Specific Entity Extraction (SEE):} Requiring retrieval of high cardinality facts such as medications and lab values or surgical procedures.
    \item \textbf{Temporal \& Longitudinal Reasoning (TLR):} Requiring tracking of changes over time, chronological event ordering or visit comparisons.
    \item \textbf{Clinical Nuance \& Status (CNS):} Questions regarding discharge instructions, patient mental status or synthesized clinical course summaries.
    \item \textbf{Absence or Negative Information (ANI):} Requiring the identification of absent information.
\end{itemize}

\begin{table}[htbp]
\begin{adjustbox}{max width=1\linewidth,center}
\small
  \centering
\begin{tabular}{l c c c}
\toprule
\textbf{Category} & \textbf{Total} & \textbf{Favored by RAG} & \textbf{Favored by FC} \\
\midrule
SEE & 18 & \textbf{12 (66.7\%)} & 6 (33.3\%) \\
TLR & 10 & 3 (30.0\%) & \textbf{7 (70.0\%)} \\
CNS & 10 & \textbf{6 (60.0\%)} & 4 (40.0\%) \\
ANI & 3 & \textbf{2 (66.7\%)} & 1 (33.3\%) \\
\midrule
\textbf{Total} & 41 & \textbf{23 (56.1\%)} & 18 (43.9\%) \\
\bottomrule
\end{tabular}
  \end{adjustbox}
  \captionsetup{justification=centering}
  \caption{Manual categorization in open-ended QA of challenging metric disagreement cases showing where RAG vs FC is more reliable.}
\label{tab:rag_vs_fc_disagreement}
\end{table}

RAG shows higher performance across the \textit{SEE} and \textit{CNS} categories, while FC leads in the \textit{TLR} one, demonstrating that while RAG effectively preserves the relevant information for specific facts and clinical nuance challenging cases, it falls behind when temporal reasoning is required disrupting continuity. A further qualitative analysis providing insights on cases where RAG or FC perform better (see Appendix \S \ref{sec:appendix.app-rag-vs-fc}) showcases findings on par with the quantitative ones, with the exception of \textit{ANI} where FC has more promising although inconsistent performance.

We further scale up the above rubric by prompting \texttt{Gemini-3-flash} with Chain-of-Thought reasoning to automatically categorize all 8K+ token instances in the multi-note reasoning EHRNoteQA dataset, computing FC versus RAG success rates via min-max normalization across LLM Correctness, NLI Entailment, and Bio F1. Human evaluation of 20 randomly sampled instances shows 85\% categorization accuracy. In Table \ref{tab:fc_vs_rag_category_performance}, we show that RAG has superior performance in the long-context categories of \textit{CNS} and \textit{ANI}, showing superior signals of consistency and factual grounding where clinical nuance, discharge status or absence of information must be identified. By contrast, FC has stronger performance in \textit{TLR}, where tracking chronological events and longitudinal changes across the full context is more reliable over note fragmenting. FC and RAG are equally good in Specific Entity Extraction \textit{SEE}. Retrieval is the better default over long patient histories, winning in \textit{CNS} and \textit{ANI} both manually and at scale, except for temporal and longitudinal reasoning question \textit{TLR}, where FC prevails.

\begin{figure}[htbp]
\centering
\includegraphics[width=1\columnwidth]{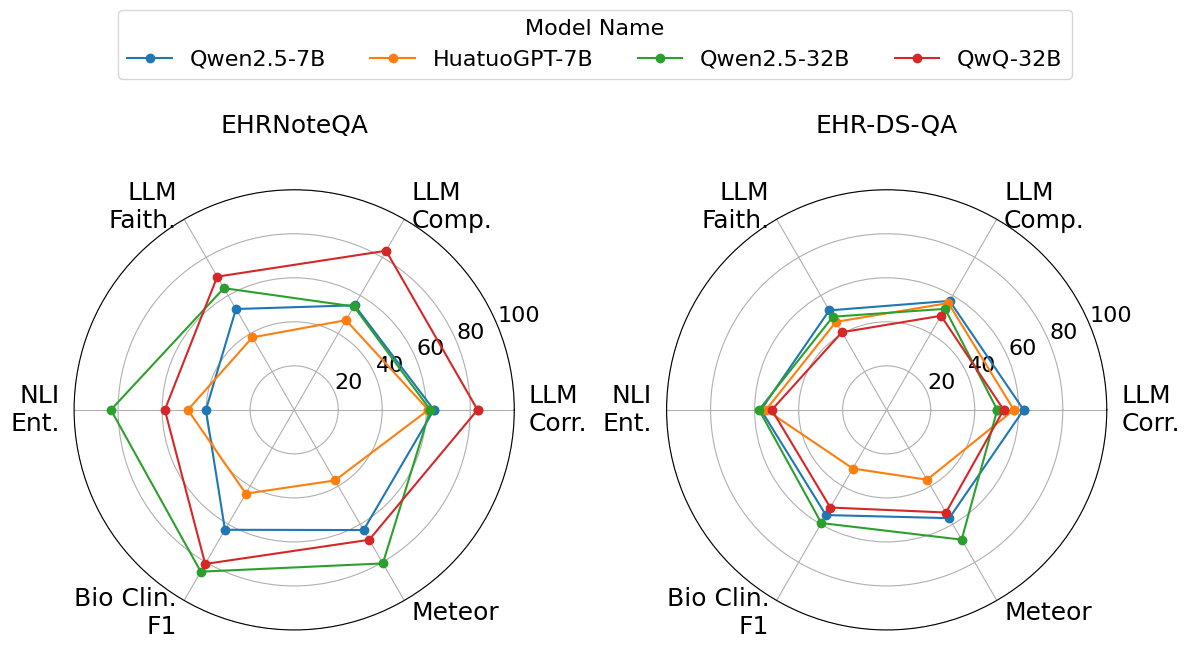}
\captionsetup{justification=centering}
\caption{Model performance across metrics for open-ended QA, normalized with min-max per metric.}
\label{fig:model_performance_metric}
\end{figure}

\begin{table}[htbp]
\begin{adjustbox}{max width=1\linewidth,center}
\small
\centering
\begin{tabular}{l r r r r r r r}
\toprule
 & & \multicolumn{2}{c}{\textbf{Favored by(\%)}} & \multicolumn{2}{c}{\textbf{FC}} & \multicolumn{2}{c}{\textbf{RAG}} \\
\cmidrule(lr){3-4} \cmidrule(lr){5-6} \cmidrule(lr){7-8}
\textbf{Category} & \textbf{Samples} & \textbf{FC} & \textbf{RAG} & \textbf{LLM Cor.} & \textbf{NLI Ent.} & \textbf{LLM Cor.} & \textbf{NLI Ent.} \\
\midrule
ANI & 4 & 25.0\% & \textbf{75.0\%} & 50.00 & 25.37 & \textbf{100.00} & \textbf{74.77} \\
CNS & 47 & 27.7\% & \textbf{61.7\%} & 67.55 & 41.94 & \textbf{72.87} & \textbf{53.58} \\
SEE & 93 & \textbf{45.2\%} & \textbf{45.2\%} & \textbf{75.54} & \textbf{61.76} & 74.73 & 56.74 \\
TLR & 41 & \textbf{53.7\%} & 36.6\% & \textbf{63.41} & \textbf{63.80} & 60.37 & 51.13 \\
\midrule
\textbf{Total} & 185 & 42.2\% & \textbf{48.1\%} & 70.27 & \textbf{56.39} & \textbf{71.62} & 55.08 \\
\bottomrule
\end{tabular}
\end{adjustbox}
\captionsetup{justification=centering}
\caption{FC vs RAG across question categories, with metric breakdown, on EHRNoteQA (8K+).}
\label{tab:fc_vs_rag_category_performance}
\end{table}

\subsection{Metrics and Insights}
\label{sec:erroranalysis}

\noindent \textbf{Metric Comparisons:} Fig.~\ref{fig:model_performance_metric} of model performance across metrics shows that instruct models lead on the single-note task (EHR-DS-QA) across semantic and NLI metrics, while larger reasoning models lead on the multi-note reasoning task (EHRNoteQA) based on LLM-as-a-judge metrics\footnote{All- and long-context metric correlations in App.\ref{sec:appendix.metric_correlation}}. Significance testing in Table~\ref{tab:sig_reasoning} confirms that the reasoning model, QwQ-32B, significantly outperforms Qwen2.5-32B on LLM Correctness for multi-note, but not on NLI or BioClinical F1, and not on any metric for the single-note. Reasoning models give factually consistent answers that go beyond the gold reference. LLM Correctness rewards this, as its rubric permits compatible additional information, whereas NLI and BioClinical F1 demand stricter reference alignment and penalize the same answers, highlighting an evaluation gap. 


\vspace{0.1em}

\noindent \textbf{Metric Insights:} A qualitative analysis of metric disagreements reveals three common failure modes: (1) models adding unsupported details, inflating Correctness over Faithfulness; (2) correct answers with low lexical overlap penalised by METEOR/BERTScore and (3) clinically aligned answers flagged by NLI due to negation or phrasing differences. These point to the need for more reasoning-appropriate metrics (\S Appendix \ref{app:appendix.metric_insights}).

\section{Conclusion}

We comprehensively evaluate LLM Long-Context (LC) comprehension on longitudinal,
patient-centric EHRs across multiple datasets, models, and metrics under both Full-Context (FC)
and RAG settings. Feeding only the most relevant content outperforms its full-notes
counterpart, showing that LLMs struggle to identify the most relevant information
in LC settings. While reasoning models gain on complex multi-note cases,
they fail to generate factually consistent and compatible answers that
go beyond the gold reference. Medically fine-tuned models do not exhibit superior
performance despite their promise. Hybrid RAG with re-ranking outperforms FC, with
chunk-based RAG benefiting single-note and hierarchical RAG benefiting complex
multi-note reasoning and smaller models benefiting most. A category-level analysis
confirms RAG’s advantage in clinical nuance and absence-of-information recognition,
while FC retains an advantage for longitudinal reasoning.


\subsection*{Limitations}

Our work focuses on MIMIC-derived datasets, which represent English-only data from a single U.S. medical center. As such, performance may not generalize to other languages, populations, and healthcare systems. Furthermore, the de-identified nature of the data limits us from examining potential cultural and other biases of LLMs over long contexts. Importantly, although our work presents an early step towards benchmarking long-context LLM performance in longitudinal patient-centric settings, our findings are intended solely for research purposes. Results reflect model performance on question answering benchmarks and should not be interpreted as guarantees of clinical safety, equitable performance, or readiness for clinical deployment.
Beyond the complexities of modeling clinical notes, real-world records comprise heterogeneous data from other sources and modalities. Our work is therefore limited in its assessment of long-context by focusing only on the textual modality, something we aim to address in future work.
Our analysis also highlights limitations in current evaluation metrics, specifically regarding the assessment of reasoning, the lack of transparency in LLM-as-a-judge frameworks, and the difficulty of accurately measuring faithfulness when benchmarks are supplemented with additional information.
Finally, while we focus on Qwen-based models due to their competitive performance, we acknowledge that an evaluation of a wider range of LLMs could offer a more complete picture of the LLM landscape in long-form medical QA. 

\subsection*{Ethics Statement}

This work uses MIMIC-III, MIMIC-IV, and four derivative datasets (EHRNoteQA, EHR-DS-QA, RadQA, CliniQG4QA) accessed under the PhysioNet Credentialed Health Data License 1.5.0. All datasets contain de-identified patient records in accordance with HIPAA Safe Harbor standards. 
We exclusively used 
open-weight LLMs to ensure no patient data were transmitted to third-party proprietary systems.

\section*{Acknowledgements}

This work was supported by the Engineering and Physical Sciences Research Council [grant number EP/Y009800/1], through funding from Responsible Ai UK (KP0016) as a Keystone project led by Maria Liakata, also supporting Talia Tseriotou and Jenny Chim. 

The authors would like to thank Prof.~Dr.~Ferdi S\"onmez for his guidance and the anonymous reviewers for their valuable feedback. Special thanks to the creators and maintainers of the MIMIC databases for their commitment to open science.




\bibliography{custom}


\appendix

\section{Inference Setup} 
\label{sec:compute}

\begin{itemize}
    \item \textbf{Inference Engine:} 
    Nvidia TensorRT.
    \item \textbf{Quantization:}  
    FP8.
    \item \textbf{Hardware:}
    7B parameter models: Deployed on single Nvidia H100 GPUs with 40GB VRAM.
    32B parameter models with extended context: Deployed on dual Nvidia H100 GPUs, each equipped with 80GB VRAM.
    A total of 260 GPU hours was spent.
    \item \textbf{Libraries:} 
    Qdrant, DSPy, Litellm, pandas, Evaluate, Fastembed, Pylate
\end{itemize}

\section{Hyperparameters}
\label{sec:hyperparams}

\begin{table}[h!]
\centering
\begin{adjustbox}{max width=\linewidth}
\begin{tabular}{@{} l l l @{}}
\toprule
\multicolumn{2}{l}{\textbf{Hyperparameter}} & \textbf{Value} \\
\midrule
\multirow{5}{*}{LLM} & temperature & inst: $0$, reas: $1$ \\
 & freq\_penalty & $0$ \\
 & pres\_penalty & $0$ \\
 & \multirow{2}{*}{think tokens} & QwQ: $20k$ \\
 & & HuatuoGPT: $8k$ \\
\midrule
\multirow{5}{*}{RAG} & top\_k & $2 \times$ num chunks \\
 & $k_{\text{RRF}}$ & $60$ (recommended by Qdrant)\\
 & $k_1$ & TF saturation \\
 & $b$ & length norm. \\
 & avgdl & avg. doc length \\
\bottomrule
\end{tabular}
\end{adjustbox}
\captionsetup{justification=centering}
\caption{Hyperparameter Configuration}
\label{tab:hyperparameters_small}
\end{table}

\section{Datasets} \label{sec:appendix.preprocessing}

\subsection{Context Data Distributions}

\begin{figure}
    \centering
    \includegraphics[width=\linewidth]{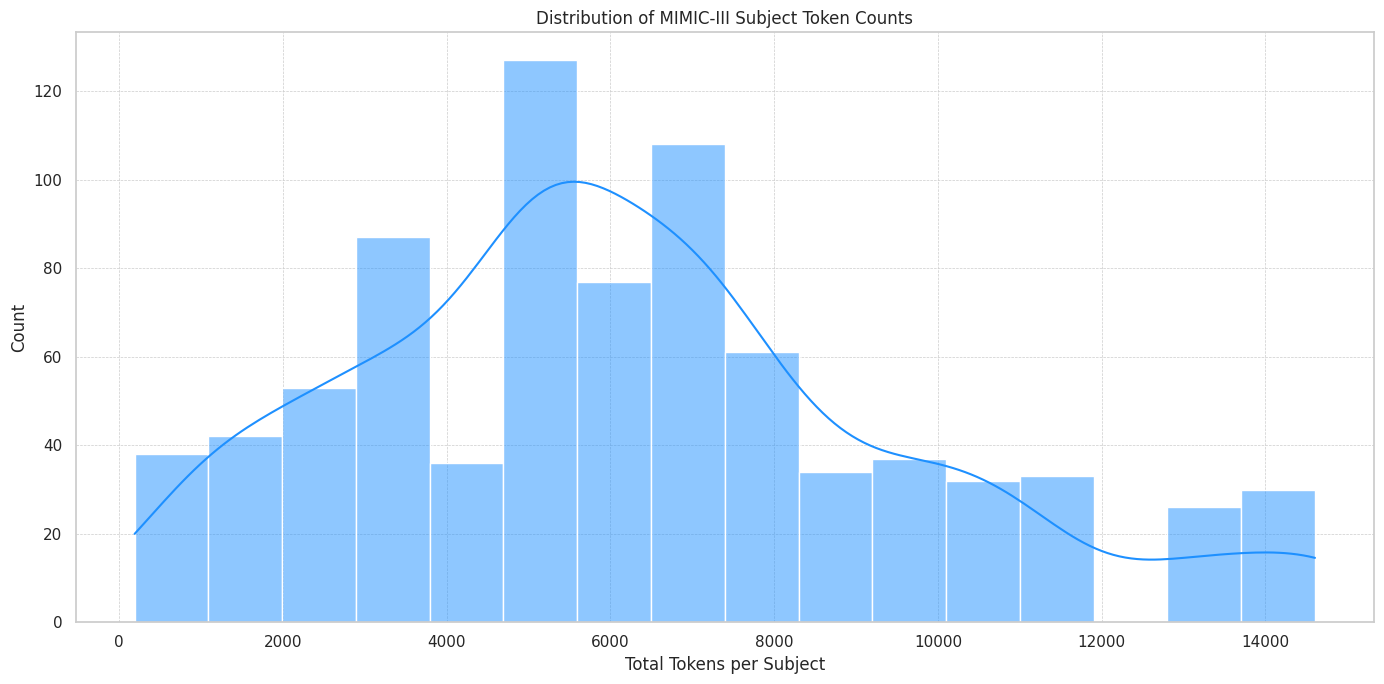}
    \caption{MIMIC-III Context Distributions}
    \label{fig:mimiciii-context}
\end{figure}

\begin{figure}
    \centering
    \includegraphics[width=\linewidth]{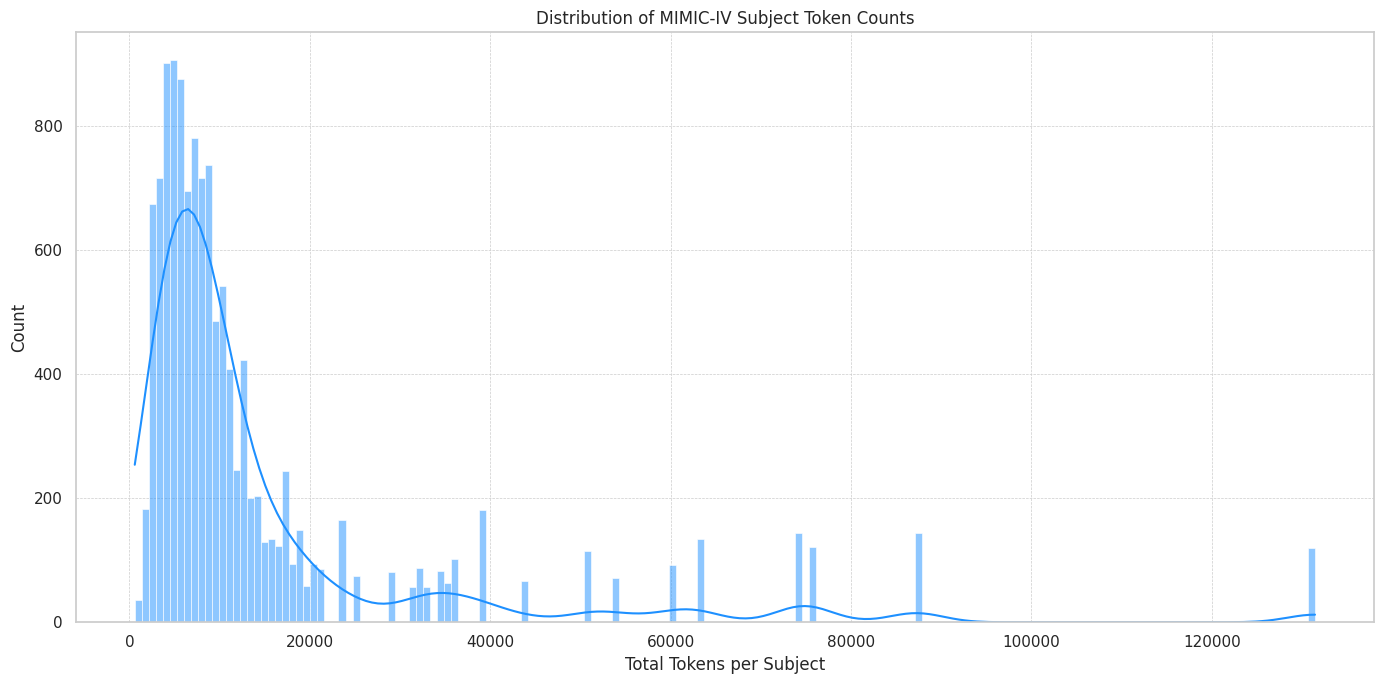}
    \caption{MIMIC-IV Context Distributions}
    \label{fig:mimiciv-context}
\end{figure}

Token-length distributions per dataset are shown in Figures~\ref{fig:mimiciii-context} and \ref{fig:mimiciv-context}.

\subsection{Context Data Preparation} \label{sec:data-prep}
We performed minimal cleaning and standardization to support retrieval and prompting, including merging notes by patient and stay, normalizing timestamps, adding note-type metadata, computing token counts for context segmentation, and filtering to human-verified subsets where applicable. 

We describe the cleaning and normalization steps applied prior to indexing and prompting:
\begin{itemize}
    \item Cleaning and de-duplication of clinical notes; removal of template boilerplate where applicable.
    \item Merging notes by patient and hospital stay; preserving note-type and section headers for downstream retrieval.
    \item Datetime standardization and chronological ordering; normalization where timezones or partial timestamps occur.
    \item Metadata augmentation (e.g., note type, encounter identifiers) to support Include DS vs Include All settings.
    \item Tokenization and token-count computation at note- and patient-level for context segmentation.
\end{itemize}

These details enable reproducibility of sampling and retrieval segmentation. 


\begin{table}[h!]
\begin{adjustbox}{max width=1\linewidth,center}
\centering 
\begin{tabular}{lcccc}
\toprule
 & \textbf{0--8K} & \textbf{8--16K} & \textbf{16--32K} & \textbf{32--128K} \\
 & \textbf{(Include All)} & \textbf{(Include All)} & \textbf{(Include All)} & \textbf{(Include All)} \\
\midrule
Include Related  & 2,247  & 4,767  & 5,149  & 28,131 \\
Exclude Related  & 1,284  & 5,712  & 13,733 & 24,472 \\
Include All      & 3,934  & 9,851  & 18,909 & 41,409 \\
\bottomrule
\end{tabular}   
\end{adjustbox}
\captionsetup{justification=centering}
\caption{Context Length per Formulation on the combined EHR-DS-QA and EHRNoteQA datasets.}
\label{tab:contextlengen}
\end{table}

\begin{figure}
    \centering
    \includegraphics[width=\linewidth]{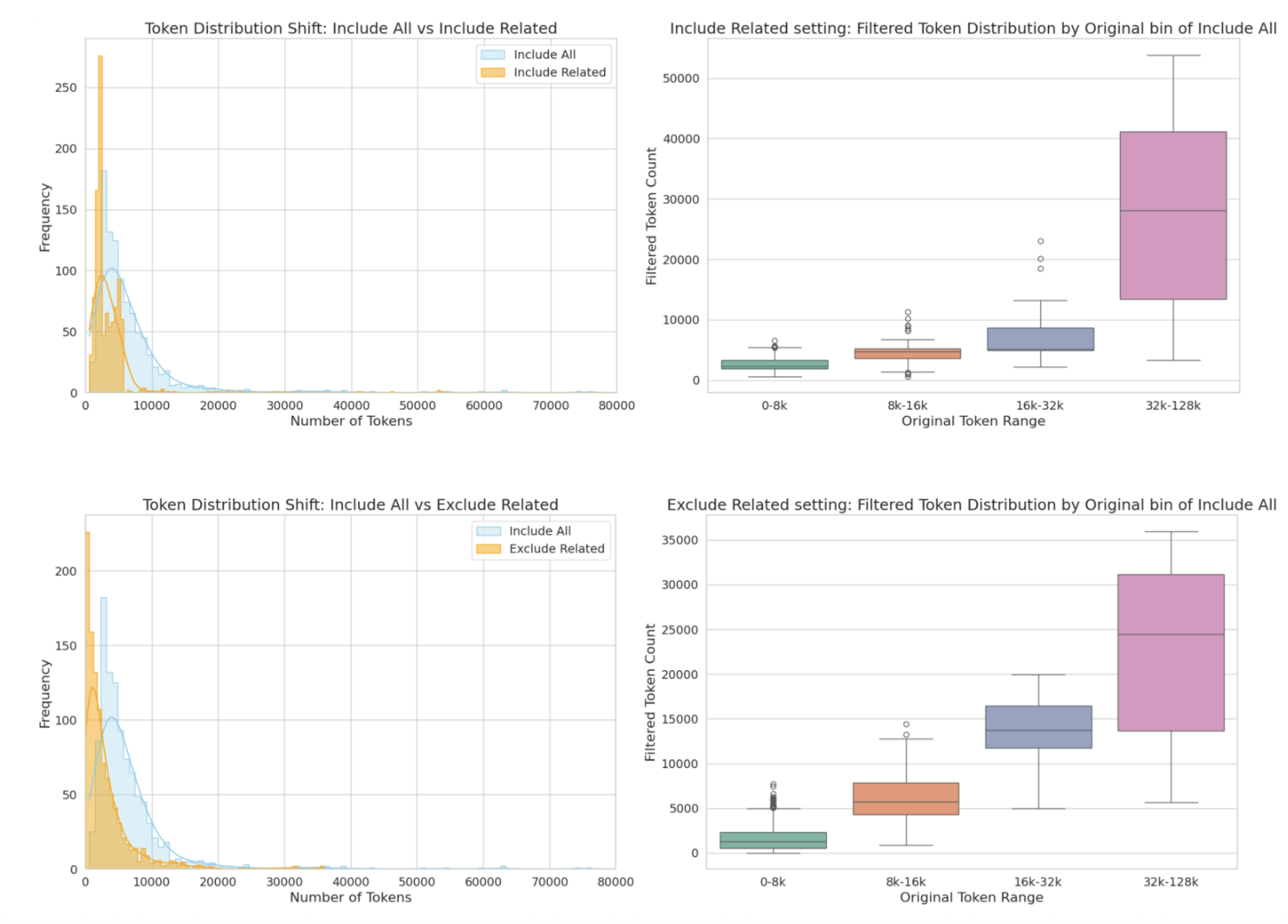}
    \caption{Distribution of tokens of Include Related versus Include All and Exclude Related versus Include All  on the combined EHR-DS-QA and EHRNoteQA datasets.}
    \label{fig:conlendistr}
\end{figure}

\begin{table}[h]
\begin{adjustbox}{max width=1\linewidth,center}
\centering
  \begin{tabular}{l l r}
    \hline
    \textbf{Dataset} & \textbf{LLM-as-a-judge} & \textbf{QA Pairs} \\
    \hline
    \multirow{3}{*}{EHR-DS-QA} & Selene-8B & 113  \\
     & Prometheus-8x7B-v2.0 & 98  \\
     & Qwen2.5-32B-Instruct & 71 \\
    \midrule
    \multirow{3}{*}{EHRNoteQA} & Selene-8B & 180 \\
     & Prometheus-8x7B-v2.0 & 220 \\
     & Qwen2.5-32B-Instruct & 42 \\
    \hline
  \end{tabular}
  \end{adjustbox}
  \captionsetup{justification=centering}
  \caption{Disagreement instances where NLI Med Contradiction is high ($>$ 0.7) and LLM Correctness is also high ($>$ 0.7).}
  \label{tab:high-contradiction}
\end{table}

\begin{table}[h]
\begin{adjustbox}{max width=1\linewidth,center}
\centering
  \begin{tabular}{l l l r}
    \hline
    \textbf{Dataset} & \textbf{LLM-as-a-judge} 1 & \textbf{LLM-as-a-judge 2} & \textbf{QA Pairs} \\
    \hline
    \multirow{3}{*}{EHR-DS-QA} & Prometheus& Qwen2.5-32B-Instruct & 520 \\
     & Prometheus & Selene-8B & 363 \\
     & Qwen2.5-32B-Instruct & Selene-8B & 
    664 \\
    \midrule
    \multirow{3}{*}{EHRNoteQA} & Prometheus & Qwen2.5-32B-Instruct & 690 \\
     & Prometheus & Selene-8B & 236 \\
     & Qwen2.5-32B-Instruct & Selene-8B & 519 \\
    \hline
  \end{tabular}
  \end{adjustbox}
  \captionsetup{justification=centering}
  \caption{Judge-pair disagreements where the absolute difference in LLM Correctness $\geq$ 0.5.}
  \label{tab:judge-disagreements}
\end{table}

\begin{table}[h!]
\begin{adjustbox}{max width=1\linewidth,center}
\centering 
\begin{tabular}{l c c} 
\toprule 
\textbf{Metric} & \textbf{Splade} & \textbf{BM25} \\ 
\midrule 
LLM Correctness & 67.30 & \textbf{69.34}  \\ 
LLM Completeness & 65.41 & \textbf{67.14}  \\ 
LLM Faithfulness & \textbf{47.64} & 46.23 \\ 
\midrule 
NLI Med Entailment & 52.68 & \textbf{54.35} \\ 
NLI Med Contradiction & 16.98 & \textbf{16.94} \\ 
\midrule 
Bio BERTScore F1 & \textbf{79.75} & 79.67 \\ 
METEOR & \textbf{43.93} & 43.63\\ 
\bottomrule 
\end{tabular}   
\end{adjustbox}
\captionsetup{justification=centering}
\caption{Comparison of Different Sparse Retrievers}
\label{tab:retrievers_comp}
\end{table}

\subsection{Context Length per Formulation}
\label{app:contextlen}

We provide the Median token context length per original Full-Context (Include All) bin based on the combined EHR-DS-QA and EHRNoteQA datasets on which we evaluate the open-ended generative QA capabilities.  
Table \ref{tab:contextlengen} shows that as expected for each of the original bins the median lengths fall compared to the \textit{Include All} setting in both the \textit{Include Related} and \textit{Exclude Related} settings, with the most drastic change being in the \textit{Include Related} setting context-bin of 16-32K tokens from a median of $\sim$19K to a median of around $\sim$5K tokens. Additionally,in Figure \ref{fig:conlendistr} we also provide the distribution of tokens of \textit{Include Related} versus \textit{Include All} and \textit{Exclude Related} versus \textit{Include All}.

\subsection{Prompt Formulations} \label{sec:prompt_specs}

We used a simple zero-shot prompt structure and tailored it explicitly for each task formulation:
\begin{itemize}
   
    \item \textbf{Extractive}: Request the model to answer by extracting the most relevant answer from the context.

    \item \textbf{Multiple-choice}: Standard multiple choice prompt.

    \item \textbf{Open-ended}: Focusing on open ended question answering favoring short, single sentence answers.

\end{itemize}

Below is the sample prompt for extractive tasks:

\begin{adjustbox}{max width=0.9\linewidth,center}
\begin{lstlisting}
System message:

Your input fields are:
1. `medical_record` (list[str]): List of patient notes (chronological order).
2. `question` (str): A question about the patient's record.
Your output fields are:
1. `answer` (str): Short single-sentence answer to the question.

All interactions will be structured in the following way, with the appropriate values filled in.

[[ ## medical_record ## ]]
{medical_record}

[[ ## question ## ]]
{question}

[[ ## answer ## ]]
{answer}

[[ ## completed ## ]]

In adhering to this structure, your objective is: 
Given a patient's medical record and a question, answer the question correctly and shortly.
\end{lstlisting}
\end{adjustbox}

\subsection{Dataset Comparisons} 
\label{app:datasets}

\definecolor{lightgray}{rgb}{0.9, 0.9, 0.9}
\definecolor{lightergray}{rgb}{0.8, 0.8, 0.8}

\begin{table*}[t]
\begin{adjustbox}{max width=1\textwidth,center}
\small
\begin{tabular}{{l}@{}{r}p{2.4cm}{r}p{1.3cm}p{1.7cm}{c}p{1.3cm}{c}p{1.6cm}{c}p{1.5cm}@{}}
\toprule
\rowcolor{lightergray}
\multicolumn{1}{c}{\textbf{Dataset}} & \multicolumn{1}{c}{\textbf{\# QA Pairs}} & \multicolumn{1}{c}{\textbf{Token Len. (max)}} & \multicolumn{1}{c}{\textbf{Task Types}} & \multicolumn{1}{c}{\textbf{Synthetic}} & \multicolumn{1}{c}{\textbf{Annotation}} & \multicolumn{1}{c}{\textbf{Reasoning}} \\
\midrule
\rowcolor{lightgray} \multicolumn{7}{|c|}{Medical Textbooks, Websites, Knowledge Bases}\\
\hline

MedQuAD & 47{,}457 & < 2K & extractive &  & automatic & $\Large\color{red}{\pmb\times}$  \\

MashQA & 34{,}808 & < 2K & extractive (multi-span) &  & experts & $\Large\color{green}{\checkmark}$ \\

MedOdyssey \tiny(En.KG) & 100 & < 128K & extrative (graph) & & model\&human & $\Large\color{green}{\checkmark}$ \\

MedQA \tiny(USMLE) & 12,723 & > 1M & MC & & experts & $\Large\color{green}{\checkmark}$ \\
\rowcolor{lightgray} \multicolumn{7}{|c|}{Biomedical Literature}\\
\hline

BioMRC \tiny(LARGE) & 812{,}707 & < 2K & MC, extractive & & automatic & $\Large\color{red}{\pmb\times}$ \\

BioASQ \tiny(b) & 5{,}729 & < 8K & extractive, generative & & experts & $\Large\color{green}{\checkmark}$  \\
PubMedQA & 1{,}000 & < 2K & MC, generative &  & experts & $\Large\color{green}{\checkmark}$ \\
\rowcolor{lightgray} \multicolumn{7}{|c|}{Clinical Case Reports}\\
\hline
CliCR & 104{,}919 & < 8K & extractive (cloze) &  & automatic & $\Large\color{green}{\checkmark}$ \\
\rowcolor{lightgray} \multicolumn{7}{|c|}{Patient Notes}\\
\hline
LongHealth & 400 & < 8K & MC & $\Large\color{gray}{\checkmark}$  & experts & $\Large\color{green}{\checkmark}$ \\
DiSCQ & 2,029 & < 2K & extractive, generative & & experts & $\Large\color{green}{\checkmark}$ \\
RadQA & 6,148 & < 16K & extractive & & experts & $\Large\color{green}{\checkmark}$ \\
CliniQG4QA & 8,824/1,287 & < 8K & extractive &  & model/experts & $\Large\color{red}{\pmb\times}$ \\
EHR-DS-QA & 156,599/478 & < 8K  & generative &  $\Large\color{gray}{\checkmark}$ & model/experts & $\Large\color{red}{\pmb\times}$ \\
EHRNoteQA & 962 & < 8K & MC, generative && experts & $\Large\color{green}{\checkmark}$ \\
\bottomrule
\end{tabular}
\end{adjustbox}
\caption{Biomedical Datasets Comparison. References for each dataset are in Table \ref{tab:citations}.}
\label{tab:dataset_landscape}
\end{table*}

Aiming to stress-test LLM capabilities under real patient-centric scenarios, our dataset selection process was based on a extensive grid of datasets. 
Table~\ref{tab:dataset_landscape} summarizes the QA datasets we analyzed in our selection process, including the EHR candidates. The chosen EHR-based, human-verified datasets provide diverse but comparable settings across generative, extractive, and MC formulations. While some exceed 8K tokens, supporting our long-context evaluation, they all provide strong augmentation potential towards a multi-note longer evaluation setting that allows content extension up to 128K for each patient (\S \ref{sec:contextform}), enabling a realistic comparison of Full-context and RAG pipelines.
We provide citations of each considered dataset from Table \ref{tab:dataset_landscape} in Table \ref{tab:citations}. 

\begin{table}[h!]
\centering
\begin{tabular}{@{} l p{4.5cm} @{}}
    \toprule
    \textbf{Dataset} & \textbf{Source} \\
    \midrule
    MedQA & \citet{jin2021disease}  \\

    MedQuAD & \citet{BenAbacha-BMC-2019}  \\
    
    MashQA &  \citet{zhu2020question}\\

    MedOdyssey & \citet{fan2024medodyssey} \\
    \hline
    \hline

    BioMRC & \citet{pappas2020biomrc} \\

    BioASQ & \citet{nentidis2025overview} \\
    PubMedQA & \citet{jin2019pubmedqa}\\
    
    \hline
    CliCR & \citet{vsuster2018clicr} \\
    
    \hline
    LongHealth & \citet{Adams2024LongHealthAQ} \\
    MediNote &  \citet{leong2024gen}\\
    DiSCQ & \citet{lehman2022learning} \\
    RadQA & \citet{Soni2022} \\
    CliniQG4QA & \citet{Yue2020CliniQG4QAGD}\\
    EHR-DS-QA & \citet{EHRDSQA2023}\\
    EHRNoteQA & \citet{Kweon2024EHRNoteQAAL} \\
    \bottomrule
\end{tabular}
\captionsetup{justification=centering}
\caption{Corresponding Citations for Datasets considered in Dataset Selection of Table \ref{tab:dataset_landscape}}
\label{tab:citations}
\end{table}

\section{Model Comparisons}

\subsection{Sparse Retrievers}

On Table \ref{tab:retrievers_comp} we provide a sparse retriever comparison on the EHR-DS-QA dataset using the Qwen2.5-7B-Instruct-1M LLM and Qwen3-Embedding-8B as the dense retriever. BM25 has better performance across LLM Correctness, LLM Completeness and NLI-based metrics while showcasing slightly inferior performance on BioClinical BERTScore F1, METEOR and LLM Faithfulness.

\subsection{Chunking Strategy}
\label{sec:chunks}

During the development phase of the RAG approach we experiment with different chunking strategies for long-context of 8K+ tokens, comparing the retrieval of 512-token chunks, 1K-token chunks with a 12-token overlap, and full notes, all indexed and retrieved with the same dense retriever. We present the results in Table \ref{tab:chunks} and note that across context size bins of 8--16K, 16--32K and 32--128K, the 512 chunking setting has the best performance across metrics with the exception of NLI metrics in the 16--32K bin.

\begin{table*}[h]
\centering
\resizebox{\textwidth}{!}{%
\begin{tabular}{lcccccccc}
\toprule
\textbf{Setting} & \textbf{Token Range} & \textbf{LLM Corr.} & \textbf{LLM Comp.} & \textbf{LLM Faith.} & \textbf{NLI Entail.} & \textbf{NLI Contra.} & \textbf{Bio Clin. F1} & \textbf{METEOR} \\
\midrule
RAG 512 & \multirow{3}{*}{8-16K} & \textbf{41.52} & \textbf{35.27} & \textbf{40.18} & \textbf{50.50} & \textbf{18.55} & \textbf{77.69} & \textbf{35.01} \\
RAG 1K          &           & 38.84 & 33.04 & 35.71 & 36.74 & 19.72 & 76.90 & 32.65 \\
RAG Full Note   &           & 39.73 & 28.57 & 36.61 & 43.71 & 19.84 & 75.49 & 33.31 \\
\midrule
RAG 512 & \multirow{3}{*}{16-32K} & \textbf{60.85} & \textbf{54.72} & \textbf{51.42} & 55.75 & 17.37 & \textbf{80.49} & \textbf{45.11} \\
RAG 1K          &            & 53.77 & 47.17 & 50.94 & \textbf{61.68} & \textbf{13.65} & 79.44 & 41.87 \\
RAG Full Note   &            & 55.19 & 47.64 & 45.28 & 60.17 & 13.66 & 78.60 & 41.80 \\
\midrule
RAG 512 & \multirow{3}{*}{32-128K} & \textbf{52.62} & \textbf{41.28} & \textbf{43.31} & \textbf{55.90} & \textbf{18.45} & \textbf{79.80} & \textbf{42.06} \\
RAG 1K          &             & 46.51 & 38.37 & 40.99 & 49.24 & 21.02 & 79.14 & 38.98 \\
RAG Full Note   &             & 50.87 & 37.79 & 41.57 & 55.29 & 22.77 & 79.10 & 41.58 \\
\bottomrule
\end{tabular}%
}
\caption{\centering RAG chunk size ablation across token ranges and metrics for the EHR-DS-QA dataset using \texttt{Qwen2.5-7B-Instruct-1M}.}
\label{tab:chunks}
\end{table*}

\subsection{LLM-as-a-judge}
\label{sec:judgecomps}

In Table \ref{tab:high-contradiction} we examine the number of QA cases for each dataset that the different Judges cause disagreement with the NLI Med Contradiction metric. We observe that Qwen2.5-32B-Instruct \footnote{https://huggingface.co/Qwen/Qwen2.5-32B-Instruct} has far less such disagreement compared to more specialized judges. Table \ref{tab:judge-disagreements} also provides direct comparisons of LLM-as-a-judge models, demonstrating cases where LLM Correctness between such pairs is more than 0.5 and therefore high. Selene-8B  \footnote{https://huggingface.co/AtlaAI/Selene-1-Mini-Llama-3.1-8B} and Prometheus-8x7B-v2.0  \footnote{https://huggingface.co/prometheus-eval/prometheus-8x7b-v2.0} comparisons show that these two models are consistently more in agreement between them while Qwen2.5-32B-Instruct shows more independent judging abilities.

\subsection{Context Size Performance}

\begin{figure*}
\centering
\includegraphics[width=0.9\linewidth]{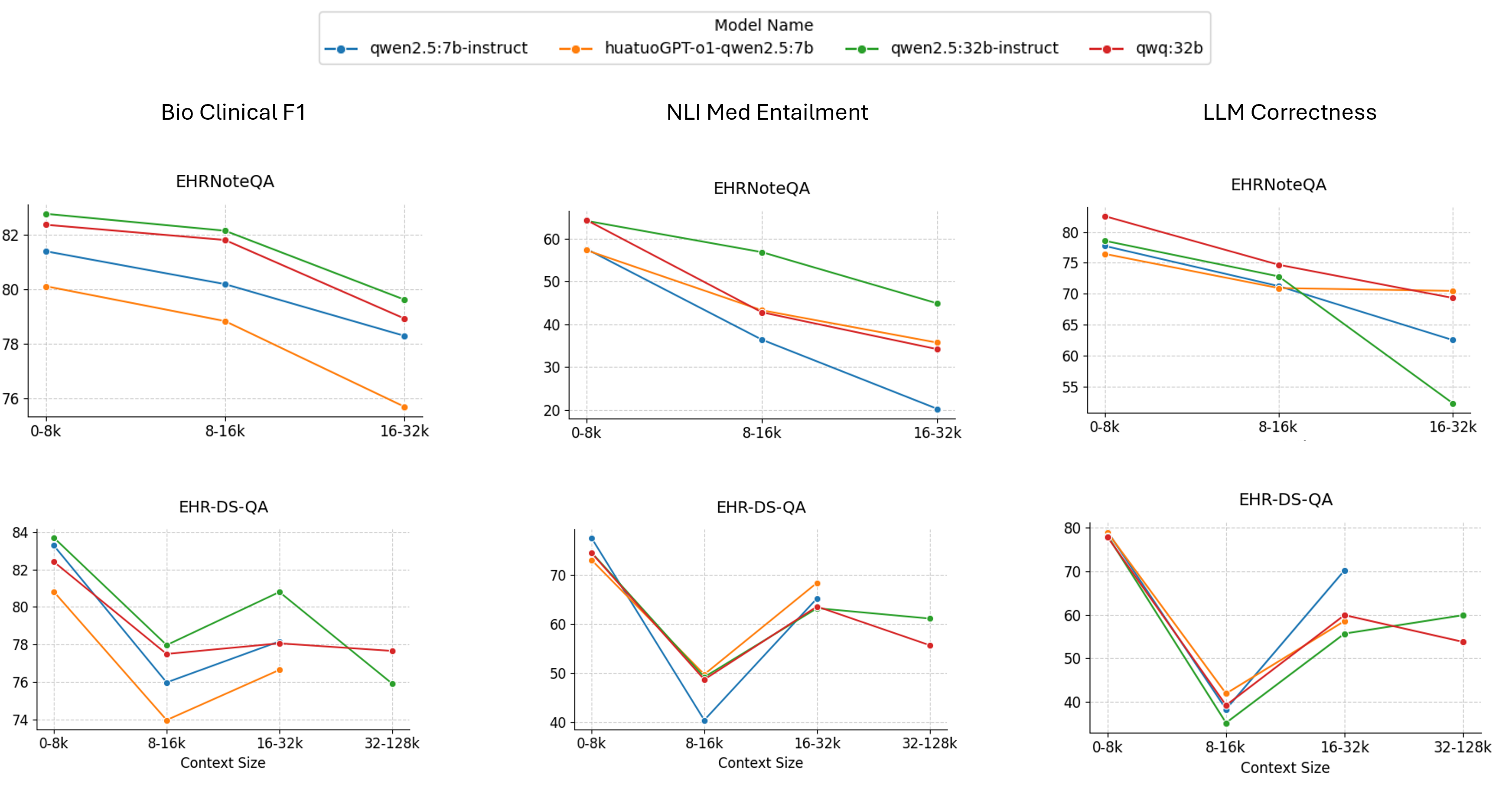}
\caption{Model performance over context size across datasets and metrics.}
\label{fig:model_performance_context}
\end{figure*}

Figure \ref{fig:model_performance_context} presents the performance of each model on open-ended generative QA across three different metrics.

\begin{figure}[htbp]
\centering
\includegraphics[width=1\columnwidth]{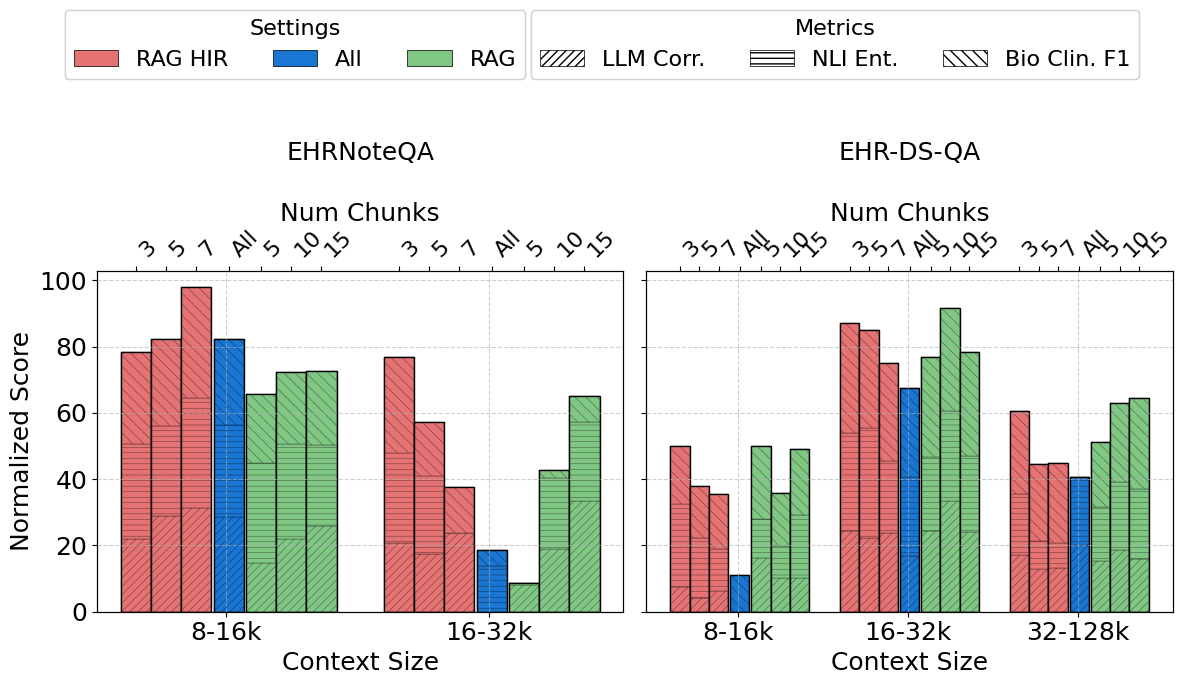}
\captionsetup{justification=centering}
\caption{RAG performance across settings with Qwen2.5:32B-Instruct, min-max normalized across LLM Correctness, NLI Entailment and Bio F1. The underlying numbers are the Qwen2.5-32B rows of Table~\ref{tab:qa_results_rag}.} 
\label{fig:rag_performance_context}
\end{figure}

\section{Statistical Significance Tests}
\label{sec:sigtests}

We performed Philipp Koehn's \citet{koehn2004statistical} paired bootstrap resampling significance
tests across experiments to examine the significance of our findings, based on 5{,}000 repetitions. 
For each comparison we bootstrap-resample questions with replacement, keeping system scores paired within each resample, and report the fraction of resamples in which the first system's mean does not exceed the second's as a one-sided $p$-value ($p<0.05$ as significant). Being paired and distribution-free, the test directly addresses whether one system is significantly better on the same questions, without normality assumptions.

In this section we present our results with a corresponding table and discussion per Claim. Claims 1--4 use the full evaluation set over all four context bins, in the \textit{Include All} setting (Claims 1 and 4), the \textit{Exclude All} setting (Claim 2) and \textit{Exclude Related} versus \textit{Exclude All} (Claim 3). Claim 5 uses the 8K+ bins of the open-ended datasets; each question is scored by the mean of LLM Correctness, NLI Entailment and BioClinical F1, and the RAG setting per model is selected based on the best model-metric average, compared against the \textit{Include All setting}.

\vspace{0.4em}
\noindent \textbf{Claim 1: Medically fine-tuning does not improve over the base model.} Strongly supported no statistically significant differences between the medically fine-tuned HuatuoGPT-o1-7B and its instruction-tuned base Qwen2.5-7B across metrics on either of the open-ended datasets: EHRNoteQA and EHR-DS-QA (Table~\ref{tab:sig_finetune}). We therefore describe this as an absence of measurable benefit rather than as a degradation.

\begin{table}[h!]
\begin{adjustbox}{max width=1\linewidth,center}
\centering
\begin{tabular}{llrrcc}
\toprule
\textbf{Dataset} & \textbf{Metric} & \multicolumn{1}{c}{\textbf{HuatuoGPT}} & \multicolumn{1}{c}{\textbf{Qwen2.5 7B}} & \textbf{$p$-value} & \textbf{$p<0.05$} \\
 & & \multicolumn{1}{c}{mean} & \multicolumn{1}{c}{mean} & & \\
\midrule
\multirow{3}{*}{EHRNoteQA} & LLM Correctness & 0.7539 & 0.7630 & 0.7936 & \small{\XSolid} \\
 & NLI Entailment & 0.5414 & 0.5294 & 0.2184 & \small{\XSolid} \\
 & BioClinical F1 & 0.7979 & 0.8111 & 1.0000 & \small{\XSolid} \\
\multirow{3}{*}{EHR-DS-QA} & LLM Correctness & 0.6993 & 0.7021 & 0.5690 & \small{\XSolid} \\
 & NLI Entailment & 0.6558 & 0.6859 & 0.8980 & \small{\XSolid} \\
 & BioClinical F1 & 0.7892 & 0.8075 & 0.9996 & \small{\XSolid} \\
\bottomrule
\end{tabular}
\end{adjustbox}
\captionsetup{justification=centering}
\caption{Paired bootstrap tests comparing the medically fine-tuned HuatuoGPT-o1-7B with the instruction-tuned base Qwen2.5-7B in the \textit{Include All} setting, across the open-ended generative datasets and metrics.}
\label{tab:sig_finetune}
\end{table}

\vspace{0.4em}
\noindent \textbf{Claim 2: Smaller models exhibit stronger memorization than larger ones.} Strongly supported that Qwen2.5 7B significantly outperforms Qwen2.5 32B in the \textit{Exclude All} setting on five of the six dataset-metric pairs, the exception being BioClinical F1 on EHR-DS-QA (Table~\ref{tab:sig_memorization}).

\begin{table}[h!]
\begin{adjustbox}{max width=1\linewidth,center}
\centering
\begin{tabular}{llrrcc}
\toprule
\textbf{Dataset} & \textbf{Metric} & \multicolumn{1}{c}{\textbf{Qwen2.5 7B}} & \multicolumn{1}{c}{\textbf{Qwen2.5 32B}} & \textbf{$p$-value} & \textbf{$p<0.05$} \\
 & & \multicolumn{1}{c}{mean} & \multicolumn{1}{c}{mean} & & \\
\midrule
\multirow{3}{*}{EHRNoteQA} & LLM Correctness & 0.0699 & 0.0133 & 0.0000 & \checkmark \\
 & NLI Entailment & 0.2242 & 0.1090 & 0.0000 & \checkmark \\
 & BioClinical F1 & 0.6784 & 0.6658 & 0.0000 & \checkmark \\
\multirow{3}{*}{EHR-DS-QA} & LLM Correctness & 0.2284 & 0.1811 & 0.0002 & \checkmark \\
 & NLI Entailment & 0.3264 & 0.1910 & 0.0000 & \checkmark \\
 & BioClinical F1 & 0.7222 & 0.7225 & 0.5440 & \small{\XSolid} \\
\bottomrule
\end{tabular}
\end{adjustbox}
\captionsetup{justification=centering}
\caption{Paired bootstrap tests on memorization, comparing Qwen2.5 7B with Qwen2.5 32B in the \textit{Exclude All} setting across open-ended generative datasets and metrics.}
\label{tab:sig_memorization}
\end{table}

\vspace{0.4em}
\noindent \textbf{Claim 3: Supplementary material is informative for generative tasks.} Strongly supported that  \textit{Exclude Related} significantly outperforms \textit{Exclude All} on five of the six dataset-metric pairs of the open-ended generative benchmarks, the exception being NLI Entailment on EHR-DS-QA (Table~\ref{tab:sig_supplementary}).

\begin{table}[h!]
\begin{adjustbox}{max width=1\linewidth,center}
\centering
\begin{tabular}{llrrcc}
\toprule
\textbf{Dataset} & \textbf{Metric} & \multicolumn{1}{c}{\textbf{Excl. Related}} & \multicolumn{1}{c}{\textbf{Excl. All}} & \textbf{$p$-value} & \textbf{$p<0.05$} \\
 & & \multicolumn{1}{c}{mean} & \multicolumn{1}{c}{mean} & & \\
\midrule
\multirow{3}{*}{EHRNoteQA} & LLM Correctness & 0.1968 & 0.0133 & 0.0000 & \checkmark \\
 & NLI Entailment & 0.1685 & 0.1090 & 0.0000 & \checkmark \\
 & BioClinical F1 & 0.7212 & 0.6658 & 0.0000 & \checkmark \\
\multirow{3}{*}{EHR-DS-QA} & LLM Correctness & 0.2187 & 0.1811 & 0.0148 & \checkmark \\
 & NLI Entailment & 0.2072 & 0.1910 & 0.1784 & \small{\XSolid} \\
 & BioClinical F1 & 0.7272 & 0.7225 & 0.0300 & \checkmark \\
\bottomrule
\end{tabular}
\end{adjustbox}
\captionsetup{justification=centering}
\caption{Paired bootstrap tests comparing \textit{Exclude Related} with \textit{Exclude All} for Qwen2.5 32B, assessing whether supplementary patient material is informative, across open-ended generative datasets and metrics.}
\label{tab:sig_supplementary}
\end{table}

\vspace{0.4em}
\noindent \textbf{Claim 4: Reasoning models offer an advantage in complex multi- versus single- note open-ended QA on LLM Correctness but not on NLI and BioClinical metrics.} 
Supported with QwQ 32B significantly outperforming Qwen2.5 32B on LLM Correctness only for EHRNoteQA (multi-note), but not on any metric on EHR-DS-QA (single-note), the single-note dataset (Table~\ref{tab:sig_reasoning}). This suggests that reasoning models generate factually consistent and compatible answers that go beyond the gold reference, which LLM Correctness rewards as it permits compatible additional information, whereas NLI and BioClinical F1 penalize the same answers because of their stricter reference alignment requirements. This further supports our argument that more reasoning-appropriate metrics are needed (\S\ref{sec:erroranalysis}).

\begin{table}[h!]
\begin{adjustbox}{max width=1\linewidth,center}
\centering
\begin{tabular}{llrrcc}
\toprule
\textbf{Dataset} & \textbf{Metric} & \multicolumn{1}{c}{\textbf{QwQ 32B}} & \multicolumn{1}{c}{\textbf{Qwen2.5 32B}} & \textbf{$p$-value} & \textbf{$p<0.05$} \\
 & & \multicolumn{1}{c}{mean} & \multicolumn{1}{c}{mean} & & \\
\midrule
\multirow{3}{*}{EHRNoteQA} & LLM Correctness & 0.8095 & 0.7701 & 0.0000 & \checkmark \\
 & NLI Entailment & 0.5972 & 0.6269 & 0.9574 & \small{\XSolid} \\
 & BioClinical F1 & 0.8218 & 0.8257 & 0.9980 & \small{\XSolid} \\
\multirow{3}{*}{EHR-DS-QA} & LLM Correctness & 0.6661 & 0.6678 & 0.5372 & \small{\XSolid} \\
 & NLI Entailment & 0.6647 & 0.6752 & 0.6888 & \small{\XSolid} \\
 & BioClinical F1 & 0.8038 & 0.8114 & 0.9706 & \small{\XSolid} \\
\bottomrule
\end{tabular}
\end{adjustbox}
\captionsetup{justification=centering}
\caption{Paired bootstrap tests comparing the QwQ 32B reasoning and Qwen2.5 32B models under the \textit{Include All} setting, across open-ended generative datasets and metrics.}
\label{tab:sig_reasoning}
\end{table}

\vspace{0.4em}
\noindent \textbf{Claim 5: RAG outperforms FC on single-note QA, with hierarchical RAG superior for multi-note reasoning in smaller models.} 
Supported through testing on 8K+ context and selecting the best RAG configuration per dataset and model by metric average against \textit{Include All} (Table~\ref{tab:sig_rag}). RAG significantly outperforms \textit{Include All} on the single-note EHR-DS-QA across all three models. For the more complex multi-note EHRNoteQA, significant gains are observed for the smaller 7B models, while Qwen2.5 32B shows no significant improvement under either open-ended or multiple-choice formulations, suggesting that larger models benefit less from retrieval in complex multi-note reasoning settings.

\begin{table}[h!]
\begin{adjustbox}{max width=1\linewidth,center}
\centering
\begin{tabular}{llcccc}
\toprule
\textbf{Dataset} & \textbf{Model/Setting} & \textbf{RAG avg} & \textbf{FC avg} & \textbf{p-value} & \textbf{p$<$0.05} \\
\midrule
\multicolumn{6}{c}{\textbf{Open-ended}} \\
\midrule
\multirow{3}{*}{EHRNoteQA} & HuatuoGPT-o1 7B (RAG HIR 5)  & 0.6863 & 0.6252 & 0.0016 & \checkmark \\
 & Qwen2.5 7B (RAG HIR 5)        & 0.6882 & 0.6023 & 0.0000 & \checkmark \\
 & Qwen2.5 32B (RAG HIR 3)       & 0.6897 & 0.6943 & 0.5998 & \small{\XSolid}  \\
\multirow{3}{*}{EHR-DS-QA} & HuatuoGPT-o1 7B (RAG 15)     & 0.7024 & 0.6193 & 0.0040 & \checkmark \\
 & Qwen2.5 7B (RAG 10)           & 0.6748 & 0.6164 & 0.0110 & \checkmark \\
 & Qwen2.5 32B (RAG 10)          & 0.6703 & 0.6275 & 0.0132 & \checkmark \\
\midrule
\multicolumn{6}{c}{\textbf{Multiple-Choice}} \\
\midrule
\multirow{3}{*}{EHRNoteQA} & HuatuoGPT-o1 7B (RAG HIR 7)  & 0.6718 & 0.6252 & 0.0056 & \checkmark \\
 & Qwen2.5 7B (RAG HIR 7)        & 0.6813 & 0.6023 & 0.0000 & \checkmark \\
 & Qwen2.5 32B (RAG HIR 7)       & 0.7132 & 0.6943 & 0.1306 & \small{\XSolid}   \\
\bottomrule
\end{tabular}
\end{adjustbox}
\captionsetup{justification=centering}
\caption{Paired bootstrap tests for the best on
average RAG strategy per dataset and model (using metric average) versus Full-Context (FC) in the \textit{Include All} setting for 8K+ context tokens.}
\label{tab:sig_rag}
\end{table}

\section{Metrics}
\subsection{Correlations}
\label{sec:appendix.metric_correlation}

\begin{figure*}
    \centering
    \includegraphics[width=0.9\linewidth]{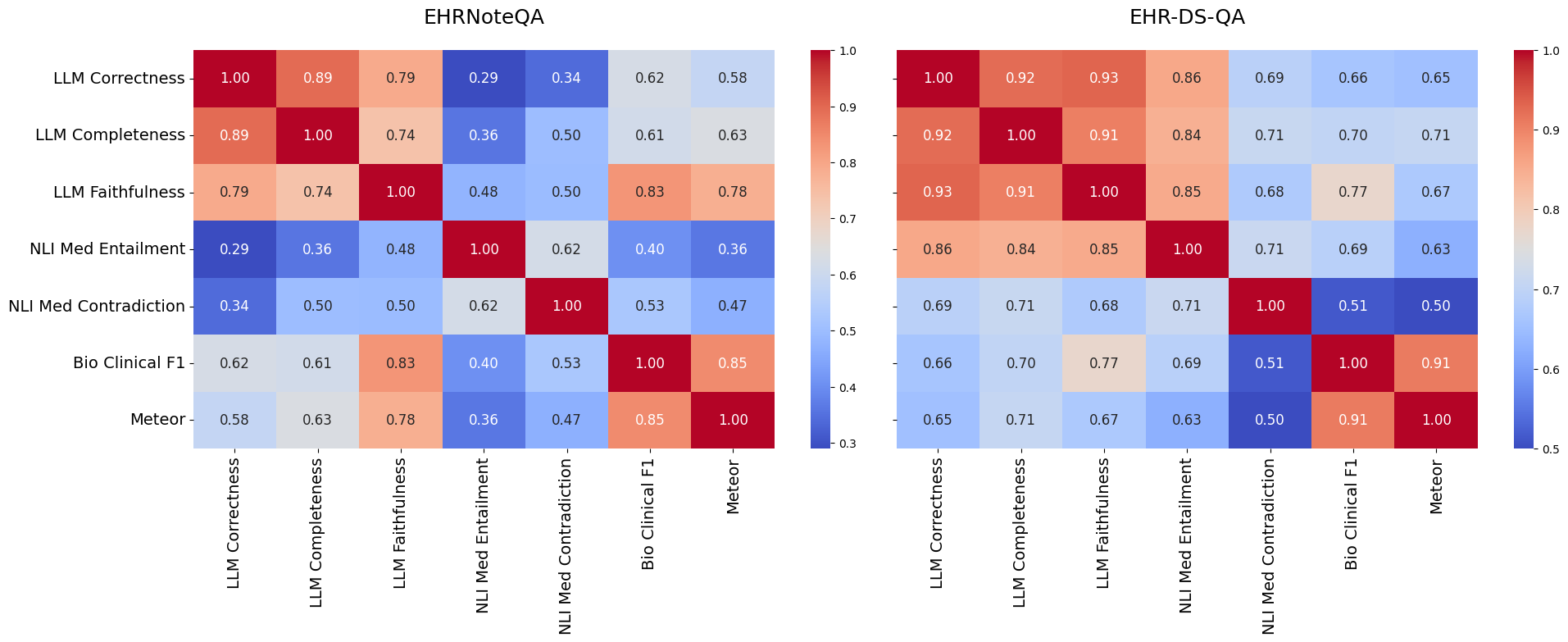}
\caption{Metric correlations across all non-exclude settings.}
\label{fig:correlation_matrix}
\end{figure*}

\begin{figure*}
    \centering
    \includegraphics[width=0.9\linewidth]{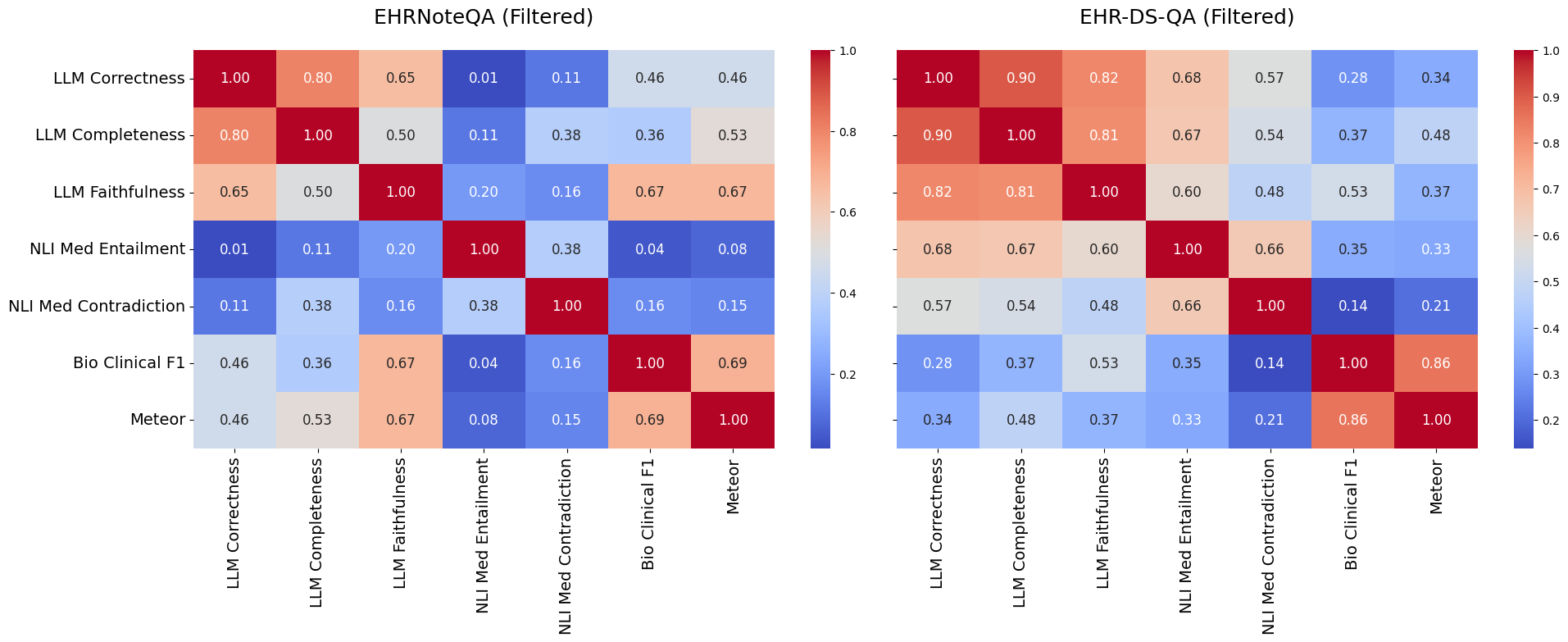}
\captionsetup{justification=centering}
\caption{Metric correlations across all non-exclude settings filtered by LLM Correctness $> 50$ and Context $>= 16$K.}
\label{fig:correlation_matrix_long}
\end{figure*}

We examine correlations between evaluation metrics to identify metric independence. On EHRNoteQA and EHR-DS-QA across all non-exclude settings (Figure~\ref{fig:correlation_matrix}) as well as across long-context high LLM Correctness non-exclude settings (Figure~\ref{fig:correlation_matrix_long}), LLM-as-a-judge evaluation metrics exhibit strong inter-correlations while METEOR and BERTScore are highly correlated. Although the above correlations remain high, they are less pronounced for the long-context high LLM Correctness setting of Figure~\ref{fig:correlation_matrix_long}. In general all metrics are less correlated for the long-context high LLM Correctness setting.

\subsection{Blind Spots}  \label{app:appendix.metric_insights}
Table \ref{tab:metrics_insights} showcases a qualitative view of what our metrics are capturing and missing, based on an analysis of disagreement on the open-ended generative QA benchmarks.

\begin{table*}[t]
\begin{adjustbox}{max width=1\textwidth,center}
\small
\begin{tabular}{@{}p{4.2cm}p{4.2cm}p{4.0cm}p{8.4cm}@{}}
\toprule
\textbf{What they capture well} & \textbf{What they miss} & \textbf{Typical pattern} & \textbf{Examples (dataset \#subject)} \\
\midrule

\multicolumn{4}{@{}l@{}}{\textbf{LLM Correctness}} \\
\midrule
Whether the main claim is right. & Unsupported add-ons that don’t change the core fact. & Correctness high; Faithfulness low (adds extra). & EHRNoteQA \#10043423 (core claim right, adds recommendation); 
\newline EHRDSQA \#10094318 (chief complaint right, extra symptom). \\

\midrule
\multicolumn{4}{@{}l@{}}{\textbf{LLM Completeness}} \\
\midrule
Coverage of the requested pieces/elements. & Penalizes omissions but not ungrounded embellishments. & Completeness high; Faithfulness low (fully covers, then embellishes). & EHRNoteQA \#10043423 (covers requested items + extras); 
\newline EHRDSQA \#19796003 (lists largely complete; minor extras elsewhere reduce faithfulness/F1).\\

\midrule
\multicolumn{4}{@{}l@{}}{\textbf{LLM Faithfulness}} \\
\midrule
Grounding to evidence/gold; flags hallucinated or embellished content. & Can be over-strict on benign context or plausible but unsupported expansions. & Correct/Complete high; Faithfulness low (ungrounded detail). & EHRNoteQA \#10043423 (extra recommendation not supported); 
\newline EHRDSQA \#10131388 (adds follow-up/psychiatric context not explicit).\\

\midrule
\multicolumn{4}{@{}l@{}}{\textbf{NLI Med Entailment / Contradiction}} \\
\midrule
Sentence-level logical relation to gold (entailed vs. contradicted). & Sensitive to negation, hedging (\emph{“low risk”} vs \emph{“no risk”}), and phrasing templates. & Correctness high; Entailment low or Contradiction high (wording/negation mismatch). & EHRNoteQA \#10494486 (mechanism phrasing hurts entailment); 
\newline EHRDSQA \#10010655 (\emph{“no risk”} vs \emph{“low acute risk”} triggers contradiction).\\

\midrule
\multicolumn{4}{@{}l@{}}{\textbf{Meteor}} \\
\midrule
Surface-form overlap, good for close paraphrases. & Under-rewards long phrases, synonyms, and re-phrasings with low lexical overlap. Incorrectly rewards inaccuracies with lexical overlap.& Correctness high; Meteor low (semantic match, low surface match). & EHRDSQA \#10076958 (concise “diffuse ischemic bowel” vs long gold narrative); \#19796003 correct single word answer but low score; 
\newline EHRNoteQA \#10404814 (correct causal link, different wording/structure).\\

\midrule
\multicolumn{4}{@{}l@{}}{\textbf{BioClinical F1}} \\
\midrule
Token/span overlap for clinical entities; good checklist signal for missing/extra items. & Under-rewards clinically acceptable reformulations (class vs. specific drug),. Incorrectly rewards inaccuracies with lexical overlap.& Correctness high; F1 moderate/low (one entity missing or formatted differently). & EHRNoteQA \#10043423 (right ideas, entity list/format differences lower F1); 
\newline EHRDSQA \#19796003 (near-exact meds but lower F1 for small misses).\\

\bottomrule
\end{tabular}
\end{adjustbox}
\captionsetup{justification=centering}
\caption{Metric blind spots: what each metric captures and misses, based on manual analysis of metric disagreement cases, with typical patterns and representative examples from EHRNoteQA and EHR-DS-QA.}
\label{tab:metrics_insights}
\end{table*}

\section{Case Studies} \label{sec:appendix.case_studies}
Here we show some sample studies of our error analysis. 
\paragraph{EHR-DS-QA: Case 10083814 (final diagnoses).}
Three discharge summaries exist across distinct admissions; each lists diagnoses. The correct target for “final diagnoses” is the most recent summary. Earlier summaries contain different diagnoses that are no longer \emph{final} at discharge, explaining prediction–gold divergence without model hallucination.

\paragraph{EHRNoteQA: Case 15877599 (cause of AKI).}
The note supports a causal chain: gastroenteritis $\rightarrow$ increased ostomy output $\rightarrow$ severe dehydration (prerenal) $\rightarrow$ acute kidney injury. The gold captures the underlying illness; a model response may describe the immediate mechanism. Both are clinically coherent parts of the same sequence.

\paragraph{EHR-DS-QA: Case 10023117 (blood pressure at discharge).}
Predictions expressed as exact values versus ranges produce different behaviors across metrics. Ranges can be faithful to notes yet fail entailment or completeness thresholds defined against a single gold value.

\section{Qualitative Analysis: RAG vs FC} \label{sec:appendix.app-rag-vs-fc}
In Table \ref{tab:rag-vs-fc-a} we summarize our insights about case categories where either RAG or FC tend to be more beneficial, providing specific use cases by open-ended generative dataset. We also provide examples along each of our main findings below:

\begin{itemize}
\item  \emph{FC performs better for questions requiring comprehensive understanding} of the entire document or when answers are located in structured sections (i.e. lists) or temporal sequences.

\noindent Examples include:
\begin{itemize}[nosep]
    \item Questions like ``What is the patient's discharge condition?'' or ``What were the patient's discharge diagnoses?''.
    \item Temporal sequence questions, such as ``What surgeries has the patient undergone and in what order?''.
\end{itemize}

\item  \emph{
RAG outperforms FC for questions that require retrieving specific details or synthesizing scattered information} across multiple parts of the document.

\noindent Examples include:
\begin{itemize}[nosep]
    \item Questions like ``What family history does the patient have?'' or ``How many tablets of dilaudid did the patient receive?''.
    \item Synthesis tasks, such as ``What were the patient's postoperative course details?''.
    \item Asking about specific dates like ``What was the outcome of the patient's colonoscopy as described in the discharge summary from the stay starting on 2113-09-30?'' or ``What was the patient's diagnosis for the hospital admission on 2154-01-28\dots'' and other examples.
\end{itemize}

\item  \emph{
RAG tends to perform better overall in datasets where specific information is required from complex or lengthy documents}.

\noindent For example:
\begin{itemize}[nosep]
    \item In the EHRNoteQA dataset, RAG consistently outperformed FC for questions needing specific details from notes or summaries, such as ``What was the outcome of the patient's colonoscopy?''
\end{itemize}

\item  \emph{There is mixed evidence suggesting FC might perform better for tasks involving inferential reasoning} or identifying the absence of information.

\noindent For example:
\begin{itemize}
    \item Questions like ``Were there any complications during the procedure?'' where RAG retrieves statements like ``No complications,'' potentially diminishing FC's advantage.
    \item Subtle inference tasks, such as ``Does the patient have any psychological issues?'' where FC occasionally performs better, though inconsistently.
\end{itemize}

\end{itemize}

\begin{table*}[t]
\begin{adjustbox}{max width=1\textwidth,center}
\small
\begin{tabular}{@{}p{1.6cm}p{1.3cm}p{4.0cm}p{10.3cm}@{}}
\toprule
\textbf{Insight} & \textbf{Favored} & \textbf{Explanation} & \textbf{Supporting Examples} \\
\midrule
Specific Fact Retrieval & RAG & RAG excels at extracting precise, well-defined medical facts (dates, medications, lab values, procedures) that are typically documented in structured sections of medical records. & \textbf{EHRNoteQA:} \newline
\textit{Subject 15036658:} Colonoscopy outcome from specific date (RAG: 0.632 vs.\ FC: 0.399) \newline
\textit{Subject 11049732:} Medication changes (RAG: 0.829 vs.\ FC: 0.352) \newline
\textit{Subject 17818938:} Surgical procedure for erectile dysfunction (RAG: 0.918 vs.\ FC: 0.479)
\newline \newline
\textbf{EHR-DS-QA:} \newline
\textit{Subject 10131388:} Dilaudid tablet count (RAG: 0.982 vs.\ FC: 0.778) \newline
\textit{Subject 19926045:} DVT medication (RAG: 0.901 vs.\ FC: 0.564) \newline
\textit{Subject 10090787:} Discharge medications (RAG: 0.425 vs.\ FC: 0.190) \\
\midrule
Temporal Information Processing & Mixed & RAG excels at explicit temporal facts (specific dates, temporal relationships) while FC is better at temporal reasoning (sequencing, duration calculation, recognizing absence of temporal information). & \textbf{RAG Advantage:} \newline
\textit{EHRNoteQA 15036658:} Specific date anchoring \newline
\textit{EHRNoteQA 18467824:} Temporal relationship between admissions \newline
\textit{EHR-DS-QA 10264949:} Nausea/vomiting timing
\newline \newline
\textbf{FC Advantage:} \newline
\textit{EHRNoteQA 11552479:} Temporal sequencing (FC: 0.696 vs.\ RAG: 0.236) \newline
\textit{EHR-DS-QA 10751849:} Duration calculation (FC: 0.541 vs.\ RAG: 0.320) \newline
\textit{EHR-DS-QA 19397212:} Absence of temporal information (FC: 0.426 vs.\ RAG: 0.180) \\
\midrule
Medical Terminology and Technical Content & RAG & RAG performs better with specialized medical terminology, complex procedures, and technical test results due to its ability to locate and interpret specific sections containing this information. & \textbf{EHRNoteQA:} \newline
\textit{Subject 17445067:} Diagnosis and surgical procedure details (RAG: 0.708 vs.\ FC: 0.328) \newline
\textit{Subject 18122852:} MRI and EMG test findings (RAG: 0.750 vs.\ FC: 0.454) \newline
\textit{Subject 16313269:} Brain mass pathological diagnosis (RAG: 0.915 vs.\ FC: 0.372)
\newline \newline
\textbf{EHR-DS-QA:} \newline
\textit{Subject 10044189:} Necrotic ulcer treatment (RAG: 0.762 vs.\ FC: 0.465) \newline
\textit{Subject 19401508:} Treatment for hyponatremia (RAG: 0.479 vs.\ FC: 0.262) \\
\midrule
Discharge Planning and Instructions & RAG & RAG performs better on Discharge information (instructions, medications, condition) that is usually well-structured in specific sections.& \textbf{EHRNoteQA:} \newline
\textit{Subject 11690633:} Discharge condition and instructions (RAG: 0.762 vs.\ FC: 0.349) \newline
\textit{Subject 11863782:} Discharge disposition and medications (RAG: 0.749 vs.\ FC: 0.327)
\newline \newline
\textbf{EHR-DS-QA:} \newline
\textit{Subject 10921250:} Discharge condition (RAG: 0.856 vs.\ FC: 0.466) \newline
\textit{Subject 10940920:} Discharge instructions (RAG: 0.650 vs.\ FC: 0.253) \newline
\textit{Subject 10064678:} Discharge instructions (RAG: 0.352 vs.\ FC: 0.094) \\
\midrule
Cause--Effect and Relationship Understanding & RAG & RAG is better at understanding relationships (symptom--procedure, medication--outcome, test result--action).& \textbf{EHRNoteQA:} \newline
\textit{Subject 13032648:} Causes of leg pain and surgical procedure (RAG: 0.730 vs.\ FC: 0.293) \newline
\textit{Subject 15748482:} Flomax usage reason and outcome (RAG: 0.819 vs.\ FC: 0.495) \newline
\textit{Subject 17436366:} Blood/urine culture results and actions (RAG: 0.718 vs.\ FC: 0.445)
\newline \newline
\textbf{EHR-DS-QA:} \newline
\textit{Subject 19926045:} Lovenox symptom management (RAG: 0.545 vs.\ FC: 0.334) \newline
\textit{Subject 19401508:} Admission cause and treatment (RAG: 0.372 vs.\ FC: 0.108) \\
\midrule
Holistic Patient Understanding & FC & FC excels when questions require synthesis of information across multiple document sections to build a complete picture of the patient’s status, multiple diagnoses.& \textbf{EHRNoteQA:} \newline
\textit{Subject 18753609:} Therapeutic interventions for leg pain (FC: 0.675 vs.\ RAG: 0.450)
\newline \newline
\textbf{EHR-DS-QA:} \newline
\textit{Subject 10262565:} Discharge condition (FC: 0.882 vs.\ RAG: 0.681) \newline
\textit{Subject 10049941:} Discharge diagnoses (FC: 0.579 vs.\ RAG: 0.188) \newline
\textit{Subject 10978236:} Discharge condition -- unusual circumstance (FC: 0.321 vs.\ RAG: 0.093) \\
\midrule
Absence or Negative Information Recognition & FC & While findings remain inconsistent with both RAG and FC demonstrating mostly low scores, FC seems more promising at recognizing when information is absent or when negative findings are documented as it can assess the entire document context. & \textbf{EHR-DS-QA:} \newline
\textit{Subject 19397212:} Absence of symptom presentation timing (FC: 0.426 vs.\ RAG: 0.180) \newline
\textit{Subject 10751849:} Absence of major procedures (FC: 0.243 vs.\ RAG: 0.037) \newline
\textit{Subject 10264949:} Absence of social factors (FC: 0.497 vs.\ RAG: 0.149) \newline
\textit{Subject 19397212:} Absence of age information (FC: 0.725 vs.\ RAG: 0.452) \\
\bottomrule
\end{tabular}
\end{adjustbox}
\captionsetup{justification=centering}
\caption{RAG vs FC Qualitative Analysis.}
\label{tab:rag-vs-fc-a}
\end{table*}

\section{LLM-as-a-Judge Prompts}
\label{sec:judge_prompts}

\lstset{basicstyle=\ttfamily\scriptsize,breaklines=true,breakatwhitespace=true,columns=fullflexible,xleftmargin=0pt,aboveskip=0.4em,belowskip=0.4em}

Below we present the three rubric prompts used for the LLM-as-a-judge evaluation (\S\ref{sec:evalframework}). Each call receives: i) \texttt{question}, ii) \texttt{generated answer} and iii) \texttt{reference answer}, and returns a free-text \texttt{reasoning} field followed by an integer \texttt{score} between $1$--$5$. Each score is then mapped to $[0,1]$ through $(s-1)/4$. We use \textsc{Qwen-2.5-32B-Instruct} with temperature $=0$ (see \S\ref{sec:judgecomps} for model selection).

\vspace{0.3em}
\noindent\textbf{Correctness Prompt}
\nopagebreak

\vspace{0.4em}
\noindent\begin{tcolorbox}[
  colback=gray!5,
  colframe=gray!40,
  arc=4pt,
  boxrule=0.5pt,
  left=10pt, right=10pt, top=8pt, bottom=8pt,
  title={\small\textit{Correctness}},
  fonttitle=\small\color{gray},
  attach boxed title to top left={yshift=-2mm, xshift=6pt},
  boxed title style={colback=white, colframe=gray!40, boxrule=0.4pt, arc=2pt}
]

\small
You are an expert medical QA evaluator. Your task is to evaluate the factual accuracy of a prediction by comparing it strictly against the ground truth reference answer.

\medskip
\noindent\textbf{Evaluation Process}

Step 1: Identify all factual claims in the prediction\\
Step 2: For each claim, check if it contradicts the ground truth\\
Step 3:  Only penalize for contradictions or factual errors, NOT for extra details

\medskip
\noindent\textbf{Scoring Rules}

- Score 5 if ALL factual claims are accurate or neutral (no contradictions with ground truth)\\
- Score 4 if most claims are correct with few minor factual inaccuracies that don't change core meaning\\
- Score 3 if roughly half the factual claims are correct \\
- Score 2 for at least a single correct factual claim\\
- Score 1 for no correct claims at all or complete failure to answer

\medskip

CRITICAL: Extra information that does not contradict ground truth = Score 5\\
CRITICAL: Missing factual information from ground truth = reduce score\\
CRITICAL: Contradicting ground truth = reduce score\\
CRITICAL: Adding neutral/compatible information = do NOT reduce score\\
CRITICAL: No correct claims at all = Score 1

\medskip
\noindent\textbf{Example:}\quad
If the ground truth states ``patient went to hospital'' and the prediction
states ``patient went to hospital on Tuesday'', this is Score 5
because adding ``on Tuesday'' does not contradict the ground truth.

\end{tcolorbox}

\vspace{0.3em}
\noindent\textbf{Completeness Prompt}
\nopagebreak

\vspace{0.4em}
\noindent\begin{tcolorbox}[
  colback=gray!5,
  colframe=gray!40,
  arc=4pt,
  boxrule=0.5pt,
  left=10pt, right=10pt, top=8pt, bottom=8pt,
  title={\small\textit{Completeness}},
  fonttitle=\small\color{gray},
  attach boxed title to top left={yshift=-2mm, xshift=6pt},
  boxed title style={colback=white, colframe=gray!40, boxrule=0.4pt, arc=2pt}
]

\small
You are an expert medical QA evaluator. Your task is to evaluate how thoroughly the prediction covers the clinically relevant elements present in the ground truth answer.

\medskip
\noindent\textbf{Evaluation Process}

Step 1: List all elements/information present in ground truth\\
Step 2: Check if each ground truth element is covered in prediction (rewording allowed)\\
Step 3:  Score based ONLY on coverage of ground truth elements

\medskip
\noindent\textbf{Scoring Rules}

- Score 5 if ALL elements from ground truth are covered in prediction\\
- Score 4 if most elements covered with minor omissions\\
- Score 3 if roughly half the elements are covered\\
- Score 2 if few elements are covered\\
- Score 1 if ALL elements from ground truth are missing

\medskip

CRITICAL: Extra information in prediction beyond ground truth = IGNORE completely\\
CRITICAL: Only count what's missing from ground truth, never penalize for extras\\
CRITICAL: Perfect rewording of ground truth content = Score 5\\
CRITICAL: All ground truth elements are missing = Score 1

\medskip
\noindent\textbf{Example:}\quad
If ground truth says ``patient has fever'' and prediction says ``patient has fever and headache'', this is Score 5 because all ground truth content (fever) is covered. The extra ``headache'' is ignored for completeness scoring.

\end{tcolorbox}

\vspace{0.3em}
\noindent\textbf{Faithfulness Prompt}
\nopagebreak

\vspace{0.4em}
\noindent\begin{tcolorbox}[
  colback=gray!5,
  colframe=gray!40,
  arc=4pt,
  boxrule=0.5pt,
  left=10pt, right=10pt, top=8pt, bottom=8pt,
  title={\small\textit{Faithfulness}},
  fonttitle=\small\color{gray},
  attach boxed title to top left={yshift=-2mm, xshift=6pt},
  boxed title style={colback=white, colframe=gray!40, boxrule=0.4pt, arc=2pt}
]

\small
You are an expert medical QA evaluator. Your task is to evaluate whether every statement in the prediction is supported by established medical knowledge or clearly inferable from the ground truth.

\medskip
\noindent\textbf{EVALUATION FOCUS:}

- Identify any unsupported claims, hallucinations, or fabricated information\\
- Evaluate support for each factual statement in prediction\\
- Ground truth serves as primary support reference
- Consider established medical knowledge as secondary support\\

\medskip
\noindent\textbf{SCORING CRITERIA (1-5):}

Score 5: All statements in prediction are well-supported by ground truth or established medical knowledge.\\
Score 4: Nearly all statements supported, with minimal harmless unsupported details.\\
Score 3: Roughly half supported and unsupported claims present.\\
Score 2: Few supported and many unsupported claims or moderate hallucinations present.\\
Score 1: No supported claims at all, significant hallucinations, or fabrications.

\medskip

CRITICAL: Examine each factual claim in prediction for support/evidence\\
CRITICAL: Claims supported by ground truth or established medical facts score well\\ 
CRITICAL: Unsupported, fabricated, or speculative claims reduce faithfulness\\
CRITICAL: If prediction says "I don't know" when ground truth is "Cannot be answered", this is faithful (score 5)\\
CRITICAL: ALL claims are supported = Score 5\\
CRITICAL: No claim is supported at all = Score 1

\medskip
\noindent\textbf{Remember:}\quad
Focus on identifying unsupported information, not completeness or style. 

\end{tcolorbox}

\label{sec:appendix}

\end{document}